\documentclass[final,5p,times,twocolumn,authoryear]{elsarticle}

\usepackage{amssymb}
\usepackage{amsmath}

\usepackage{siunitx}
\usepackage{textgreek}
\usepackage[colorlinks=true]{hyperref}
\usepackage{url}
\usepackage{booktabs}
\usepackage{multirow}
\usepackage[flushleft]{threeparttable}
\usepackage{array}
\usepackage{tabularx}      
\usepackage{makecell}      
\usepackage[table,xcdraw]{xcolor} 
\usepackage[normalem]{ulem}

\usepackage{lineno}
\usepackage{ulem}

\journal{Medical Image Analysis}

\newif\ifreview
\reviewfalse    

\ifreview
  \newcommand{\reviadd}[1]{\textcolor{red}{#1}}
  \newcommand{\reviiadd}[1]{\textcolor{magenta}{#1}}
  \newcommand{\revidel}[1]{\textcolor{red}{\sout{#1}}}
\else
  \newcommand{\reviadd}[1]{#1}
  \newcommand{\reviiadd}[1]{#1}
  \newcommand{\revidel}[1]{}
\fi

\begin{document}

\begin{frontmatter}




\title{NeuroCBIR: A Fast and Accurate Image Retrieval System for Whole-Brain and Region-Specific MRI} 


\author[1]{Félix Nieto-del-Amor\corref{cor1}}
\author[2,1]{Jingru Fu}
\author[3]{J.-Sebastian Muehlboeck}
\author[3]{Eric Westman}
\author[3,4]{Daniel Ferreira}
\author[1,3]{Rodrigo Moreno\corref{cor1}}
\author[adninote]{\small for~the~Alzheimer’s Disease Neuroimaging Initiative}
\author[aiblnote]{\small and~the~Australian Imaging Biomarkers and Lifestyle flagship study of ageing}

\affiliation[1]{%
    organization={Division of Biomedical Imaging, KTH Royal Institute of Technology},
    city={Huddinge},
    country={Sweden}
}

\affiliation[2]{organization={Athinoula A. Martinos Center for Biomedical Imaging, Massachusetts General Hospital and Harvard Medical School},
            state={MA},
            country={USA}}

\affiliation[3]{%
    organization={Division of Clinical Geriatrics, Center for Alzheimer Research, Department of Neurobiology, Care Sciences and Society, Karolinska Institute},
    city={Stockholm},
    country={Sweden}
}

\affiliation[4]{%
    organization={Facultad de Ciencias de la Salud, Universidad Fernando Pessoa Canarias},
    city={Santa María de Guía},
    state={Las Palmas},
    country={Spain}
}

\cortext[cor1]{Corresponding authors: fenda@kth.se, rodmore@kth.se}

\fntext[adninote]{%
Data used in preparation of this article were obtained from the Alzheimer’s Disease Neuroimaging Initiative (ADNI) database      (\url{http://adni.loni.usc.edu}). As such, the investigators within the ADNI contributed to the design and implementation of ADNI and/or provided data but did not participate in analysis or writing of this report. A complete listing of ADNI investigators can be found at:       \url{http://adni.loni.usc.edu/wp-content/uploads/how_to_apply/ADNI_Acknowledgement_List.pdf}.}

\fntext[aiblnote]{%
Data used in the preparation of this article was obtained from the Australian Imaging Biomarkers 
and Lifestyle flagship study of ageing (AIBL) funded by the Commonwealth Scientific and Industrial 
Research Organisation (CSIRO) and made available at the ADNI database (\url{www.loni.usc.edu/ADNI}). 
The AIBL researchers contributed data but did not participate in analysis or writing of this report. 
AIBL researchers are listed at \url{www.aibl.csiro.au}.}

\begin{abstract}
Content-based image retrieval (CBIR) in neuroimaging enables the identification of structurally similar brain scans, supporting diagnosis, prognosis, and treatment planning; however, existing methods are often limited to small datasets, single brain regions, or coarse class labels, thereby restricting their clinical utility and generalizability. 
Here, we present NeuroCBIR, a framework for fast and flexible retrieval of both whole-brain and region-specific 3D T1w MRI scans. A total of 103 cortical and subcortical regions are extracted to enable both whole-brain and region-level queries.
NeuroCBIR leverages latent representations learned by a variational autoencoder (VAE) combined with contrastive learning, producing scan-specific embeddings that capture anatomical patterns. These embeddings were evaluated for subject re-identification, zero-shot age prediction, and zero-shot multi-class pathology stratification. Re-identification performance was high across both whole-brain and brain-region levels (mean average precision across the top-5 retrieved images (mAP@5) $\ge 98.4\%$), with robust generalization across datasets and acquisition conditions. While NeuroCBIR is not trained for age prediction or pathology stratification, zero-shot evaluations for these two tasks demonstrate that the embeddings encode meaningful information for downstream tasks.
Embedding extraction on a 4-core CPU required $\sim$18.7s per scan, whereas similarity search was effectively instantaneous (\textless0.01s).
NeuroCBIR is publicly available for brain MRI with \textgreater \num[round-mode=places,round-precision=0]{26000} precomputed T1w MRI embeddings. It supports reproducible research, region-specific flexibility, and clinically meaningful personalized diagnostic support. The software is available at \url{https://github.com/minnelab/NeuroCBIR}.
\end{abstract}

\begin{graphicalabstract}
\includegraphics{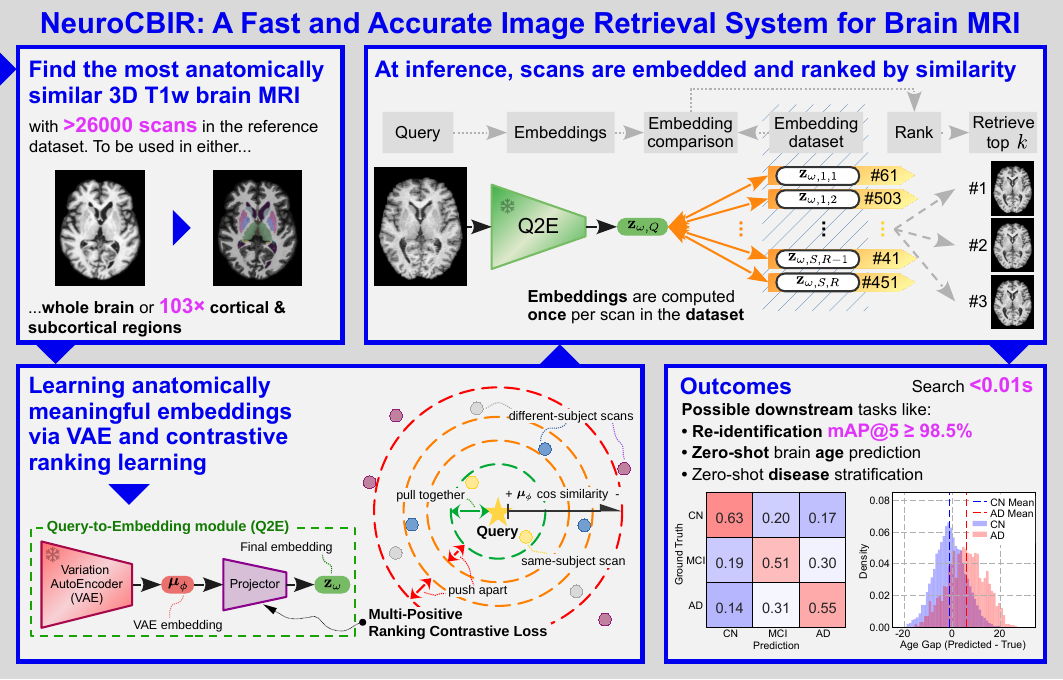}
\end{graphicalabstract}

\begin{highlights}
\item Fast whole-brain and region-level retrieval of 3D T1w MRI scans.
\item Retrieval across 103 brain regions enables flexible anatomical search.
\item State-of-the-art performance for MRI scan re-identification.
\item Embeddings support zero-shot analyses without task-specific training.
\item Code, pretrained models, and MRI embeddings are publicly released.

\end{highlights}


\end{frontmatter}



\section{Introduction}

Magnetic resonance imaging (MRI) plays a central role in the study and diagnosis of neurological disorders, providing high-resolution, non-invasive insights into brain structure, function, and pathology. With the increasing availability of large-scale neuroimaging datasets, such as those developed for the study of Alzheimer’s disease (AD) and normal aging \citep{jack_alzheimers_2008, ellis_australian_2009, oasis3}, the volume of data available to clinicians and researchers has increased greatly. Manual searches of extensive databases are neither efficient nor practical. The adoption of innovative computer-aided diagnosis methods, particularly Content-Based Image Retrieval (CBIR) \citep{huang_retrieval_2012, Onga2019, dogan_content-based_2025, tobari_domain-invariant_2025}, unlocks new possibilities. CBIR systems aim to retrieve images similar to a given query based on intrinsic image features rather than metadata. In the context of neuroimaging, CBIR offers the potential to assist clinicians and scientists by providing fast access to comparable cases, thus facilitating clinical decision making, supporting differential diagnosis, and enabling population-level analyses by efficiently identifying subjects with similar structural or functional brain patterns across large datasets.
{
\newcolumntype{Y}{>{\centering\arraybackslash}m{3cm}}
\setlength{\tabcolsep}{1.4mm}
\begin{table*}[!ht]
\begin{threeparttable}
\caption{Summary of recent literature on brain MRI retrieval and classification.}
\label{tab:lit_rev}
\centering
\small
\begin{tabularx}{\textwidth}{m{3.3cm} m{3.8cm} m{3cm} m{3.8cm} m{3.8cm}}
\toprule
\multicolumn{1}{l}{\textbf{Reference}} & 
\multicolumn{1}{l}{\textbf{Dataset / Task}} & 
\multicolumn{1}{l}{\textbf{Method}} & 
\multicolumn{1}{l}{\textbf{Evaluation}} & 
\multicolumn{1}{l}{\textbf{Performance}} \\
\midrule
\makecell[l]{\citet{huang_retrieval_2012}} & 2D T1w MRI \newline \scriptsize{(glioma, meningioma, pituitary)} & Bag-of-visual-words \newline+ metric learning & Compared tumor types \newline \scriptsize{Hit = retrieved same tumor class.} & \footnotesize{Prec@10 $\approx94\%$} \newline mAP $\approx 91\%$ \\ \midrule

\makecell[l]{\citet{Onga2019}} & 3D T1w MRI \newline ADNI2 \newline \scriptsize{AD / EMCI / LMCI / SMCI / CN} & CAE \newline+ metric learning &  \footnotesize{Clustering and classification across diagnostic groups} & \footnotesize{Compression = 4096:1} \newline Acc $\approx 81.5\%$ \\ \midrule

\makecell[l]{\footnotesize{\citet{deepak_retrieval_2020}}} & 2D T1 MRI \newline Figshare \footnotesize{(brain tumor dataset)} & GoogLeNet CNN \newline+ Siamese network & Retrieval \newline \scriptsize{Hit = retrieved same tumor class} & \footnotesize{Compression = \num[round-mode=places,round-precision=0]{77264}:1} \newline \footnotesize{Prec@10 $=97.7\%$} \newline mAP@10 $=97.6\%$ \\ \midrule

\makecell[l]{\citet{nishimaki_loc-vae_2022}} & 3D T1w MRI \newline ADNI2 (AD vs. CN) & Localized $\beta$-VAE & Classification across diagnostic groups & \footnotesize{Compression = 4096:1} \newline AUC $=0.718$ \\ \midrule

\makecell[l]{\citet{puglisi_deepbrainprint_2024}} & 2D T1w MRI \newline ADNI & DeepBrainPrint & Retrieval \newline \scriptsize{Hit = scans from same subject} & mAP@3 $=95.5\%$ \\ \midrule

\makecell[l]{\citet{muraki_isometric_2024}} & 3D T1w MRI \newline ADNI (AD vs. CN) & Isometric VAE & Classification across diagnostic groups & \footnotesize{Compression = 512:1} \newline Acc $=91.5\%$ \\ \midrule

\makecell[l]{\citet{dogan_content-based_2025}} & 2D T1 MRI \newline Multi-disease 3D brain MRI \newline AD, MS, glioma, etc. & EfficientNet-B0  \newline+ LBP + mRMR & Classification across diagnostic groups & \footnotesize{Compression = 167:1} \newline Acc $=98.9\%$ \\ \midrule

\makecell[l]{\citet{tobari_domain-invariant_2025}} & 3D T1w MRI \newline ADNI1–3, OASIS1–4, PPMI \newline AD vs. CN & SE-ADA & Classification across diagnostic groups & F1-score $=74.7\%$ \\ \midrule

\makecell[l]{\citet{denner_leveraging_2025}} & 2D radiology images \newline \scriptsize{RadImageNet, NIH14, MIMIC, CheXpert, 1.6M images, 161 pathologies} & \footnotesize{Foundation models (BiomedCLIP, CLIP, ViT, etc.), embeddings} & Retrieval task \newline Single-label class match & Prec@10 = 58.3\% \\
\bottomrule
\end{tabularx}
    \begin{tablenotes}
        {\footnotesize
        \setlength{\leftmargin}{0pt}
        \setlength{\labelsep}{0pt}
        \item[] Note: The abbreviations used in the table are as follows:
        2D/3D T1w MRI – two-/three-dimensional T1-weighted magnetic resonance imaging;
        AD – Alzheimer's disease;
        Acc – accuracy;
        AUC – area under the receiver operating characteristic curve;
        $\beta$-VAE – beta variational autoencoder;
        CAE – convolutional autoencoder;
        CN – cognitively normal;
        CNN – convolutional neural network;
        EMCI – early mild cognitive impairment;
        LBP – local binary patterns;
        LMCI – late mild cognitive impairment;
        mAP – mean average precision;
        mRMR – minimum redundancy maximum relevance feature selection;
        PCA – principal component analysis;
        PLS-DA – partial least squares discriminant analysis;
        SE-ADA – style encoder adversarial domain adaptation;
        SMCI – stable mild cognitive impairment.}
    \end{tablenotes}
\end{threeparttable}
\end{table*}
}

In Table~\ref{tab:lit_rev}, we summarize the different CBIR systems for brain MRI data found in the literature. Early CBIR systems relied on handcrafted feature extraction and statistical models. In these early approaches, features were neuroanatomically meaningful (volume or intensity of atlas regions) and retrieval performance was high but typically evaluated on specific disorders. For instance, \citet{huang_retrieval_2012} used a region-specific bag-of-visual-words model on contrast-enhanced T1 brain MRIs (meningioma, glioma, pituitary tumors), learning a similarity metric via distance metric learning. This achieved a mean average precision (mAP) $\approx 91$\% across multiple views.

In the past 5 years, deep learning has reshaped the field of CBIR. Many available methods use convolutional networks or autoencoders to learn low-dimensional embeddings directly from images. In the work of \citet{deepak_retrieval_2020}, T1-weighted tumor MRIs were passed through a pretrained convolutional neural network (CNN) and then a 2D siamese embedding trained with contrastive loss, achieving a tumor-retrieval performance of mAP@10 $\approx 97.6$\% (mAP across the top-10 retrieved images). Likewise, the study by \citet{Onga2019} combined a 3D convolutional autoencoder (3D-CAE) with deep metric learning in patients with AD. It compressed full-resolution 3D MRIs by 4096:1 while preserving AD-related features. The learned embedding improved the separation between AD and cognitively normal (CN) cases by 29.1\% compared to a vanilla 3D-CAE. Similarly, \citet{nishimaki_loc-vae_2022} proposed Loc-VAE, a localized \textbeta-VAE that enforces each latent dimension to correspond to a specific brain region. Loc-VAE compressed at 4096:1, preserving disease signals while enhancing “locality” of features. Furthermore, the study by \citet{muraki_isometric_2024} described IE-CBIR, an isometric VAE embedding with 4096:1 compression. This method achieved an accuracy of 91.5\% for discriminating AD vs. CN using the latent representation. In a different line of work, \citet{dogan_content-based_2025} combined deep and handcrafted features using a CNN (EfficientNet-B0) and Local Binary Patterns (LBP), followed by minimum redundancy–maximum relevance (mRMR) feature selection. This hybrid CBIR-classifier achieved 98.9\% accuracy over eight brain-MRI categories (including glioma, meningioma and AD). More recently, \citet{tobari_domain-invariant_2025} introduced a Style Encoder Adversarial Domain Adaptation (SE-ADA) model to improve CBIR robustness across multi-cohort MRI datasets, achieving an F1-score of 74.7\% in AD vs. CN retrieval.

Alternatively, the study by \citet{puglisi_deepbrainprint_2024} addressed subject-level re-identification\footnote{Re-identification refers to the task of correctly matching scans from the same subject across different time points or datasets, ensuring that repeated imaging of an individual can be reliably retrieved.} by generating a fingerprint from 2D T1-weighted MRIs, achieving mAP@3 of 95.5\% on the ADNI dataset. While demonstrating the feasibility of retrieving previous scans of the same patient, their study was limited to small 2D datasets.

Parallel to these MRI-focused efforts, \citet{denner_leveraging_2025} investigated the use of large-scale vision foundation models for medical CBIR in radiology. Using over 1.6 million images spanning multiple modalities and 161 pathologies, they benchmarked off-the-shelf embeddings from BiomedCLIP, CLIP, and other vision foundation models. Their results showed that foundation models can achieve competitive retrieval performance, Precision@10 = 58.3\%, highlighting the potential of general-purpose embeddings for scaling CBIR to heterogeneous radiology datasets. 

The existing CBIR approaches described above are typically constrained by limited datasets, restricted applicability to specific brain regions, and a lack of evaluation across heterogeneous MRI acquisition protocols. Moreover, to the best of our knowledge, no publicly available system currently provides a CBIR framework for multiple brain-region MRI data.

Our work leverages five large-scale datasets (ADNI, OASIS3, AIBL, MIRIAD, and SLIM), enabling evaluation across diverse populations, scanner manufacturers, and field strengths. A key advantage of our proposed CBIR system (NeuroCBIR) is that, although it employs deep learning for image retrieval, the retrieved results consist exclusively of expert-curated, real-world clinical data. 
\revidel{This design ensures that any subsequent decisions are grounded in transparent, interpretable evidence rather than in black-box model outputs.} 
We extend CBIR beyond the whole-brain level by supporting retrieval at the brain-region level, covering a total of 103 cortical and subcortical brain regions. This regional flexibility enhances medical usability by allowing clinicians and researchers to query specific regions of interest according to their diagnostic needs, thereby enabling more precise and context-driven retrieval.

From a technical perspective, we introduce a retrieval framework that couples VAE-derived latent embeddings with contrastive learning using a multi-positive ranking loss. This design exploits multiple valid matches per query, producing stable and scalable embeddings that support efficient and anatomically meaningful MRI retrieval \citep{cheng_breaking_nodate, tang_ranking-based_2023}.

A central goal of NeuroCBIR is accurate re-identification of same-subject scans at both whole-brain and region-specific levels. Simultaneously, the embedding space is explicitly structured to rank anatomically similar inter-subject samples close to each query while separating dissimilar ones, ensuring that retrieved cases are truly comparable. This provides a reliable foundation for downstream zero-shot analyses and clinical interpretation, and enables efficient re-identification within large-scale imaging repositories.

Accordingly, the main contributions of this study are:  
\begin{enumerate}
    \item Design a fast and flexible framework for both whole-brain and region-specific retrieval, thereby enhancing clinical applicability.  
    \item Evaluate retrieval robustness across multiple datasets, MRI acquisition protocols, and imaging domains (scanner manufacturer and field strength).  
    \item Evaluate whether the learned retrieval embeddings capture clinically meaningful information beyond accurate re-identification, by assessing their utility in zero-shot downstream tasks such as age estimation and disease risk inference.
    \item Publicly release a CBIR system for brain MRI, capable of retrieving the most similar real-world imaging cases to support personalized diagnosis and tailored treatment decisions. We intend to share the source code and precomputed image features, ensuring reproducibility while respecting data privacy and institutional data-sharing constraints.
\end{enumerate}

\section{Materials and Methods}

This section describes the MRI datasets and the methodologies employed in this study. Two CBIR frameworks were developed, one based on whole-brain scans and the other on segmented brain-regions. Since both frameworks share the majority of their processing pipeline, the common components are presented jointly, while the specific differences are clearly indicated throughout the text. 

All source code, trained models, and precomputed embeddings are publicly available at \url{https://github.com/minnelab/NeuroCBIR}. A reproducibility script (\texttt{dev/run\_reproducibility\_check.py}) enables full replication of model training, embedding extraction, and retrieval evaluation. This ensures that all results reported in the manuscript can be independently reproduced by other researchers.

We first introduce the datasets and their corresponding preprocessing steps. Next, we describe the methodological framework, which includes: the architecture of the variational autoencoder (VAE) and its training procedure, as well as the encoder–projector model architecture and its training using contrastive learning. Finally, we outline the evaluation metrics applied in the different experimental scenarios, together with the statistical tests used for performance assessment.

\subsection{Dataset description}

{
\sisetup{
  round-mode=places,
  round-precision=0,
  zero-decimal-to-integer=true  
}
\setlength{\tabcolsep}{4pt}
\begin{table*}[t]
\begin{threeparttable}
\caption{Summary of the MRI datasets used in this study.}
\label{tab:dem_des}
\centering
\small
\begin{tabular*}{\textwidth}
{@{\extracolsep{\fill}}l
S[table-format=5.0]
S[table-format=5.0]
S[table-format=3.1,round-mode=places,round-precision=1]
S[table-format=3.1,round-mode=places,round-precision=1]
>{\raggedleft\arraybackslash}p{1cm}
S[table-format=4.0]
S[table-format=5.0]
S[table-format=4.0]
S[table-format=5.0]
S[table-format=5.0]
S[table-format=5.0]
S[table-format=5.0]
S[table-format=5.0]}
\toprule
\multirow{2}{*}{\textbf{Dataset}} & 
\multicolumn{1}{c}{\multirow{2}{*}{\textbf{$\mathbf{N_I}$}}} & 
\multicolumn{1}{c}{\multirow{2}{*}{\textbf{$\mathbf{N_S}$}}} & 
\multicolumn{1}{c}{\multirow{2}{*}{\textbf{$\mathbf{N_R}$}}} & 
\multicolumn{1}{c}{\multirow{2}{*}{\textbf{$\mathbf{N_{T}}$}}} & 
\multicolumn{1}{c}{\multirow{2}{*}{\textbf{Age}}} & 
\multicolumn{3}{c}{\textbf{Class}} & 
\multicolumn{2}{c}{\textbf{Field strength}} & 
\multicolumn{3}{c}{\textbf{Manufacturer}} \\

\cmidrule{7-9} \cmidrule{10-11} \cmidrule{12-14}
 &   &  &  &  & & \textbf{CN} & \textbf{MCI} & \textbf{AD} & \textbf{1.5T}         & \textbf{3T}         & \textbf{SH} & \textbf{GE} & \textbf{PH}    \\

 \midrule 
ADNI & 20367 & 2389 & 2.0 & 4.1 & 75 ± 7 & 6260 & 11572 & 2524 & 7951 & 8779 & 7453 & 3563 & 1794 \\
OASIS3 & 2642 & 1303 & 1.0 & 2.0 & 71 ± 9 & 2149 & 392 & 100 & 208 & 2429 & 2642 & 0 & 0 \\
AIBL & 1276 & 685 & 1.0 & 1.8 & 74 ± 7 & 938 & 179 & 152 & 266 & 1010 & 1276 & 0 & 0 \\
MIRIAD & 706 & 69 & 1.3 & 7.6 & 70 ± 7 & 243 & 0 & 463 & 706 & 0 & 0 & 706 & 0 \\
SLIM & 1015 & 571 & 1.0 & 1.8 & 21 ± 1 & 1015 & 0 & 0  & 0 & 1015 & 1015 & 0 & 0 \\
 \bottomrule   
\end{tabular*}
\begin{tablenotes}
        {\footnotesize
        \setlength{\leftmargin}{0pt}
        \setlength{\labelsep}{0pt}
        \item[] Note: The abbreviations used in the table are as follows:
        $\mathrm{N_I}$: number of scans; 
        $\mathrm{N_S}$: number of subjects; 
        $\mathrm{N_R}$: average repeated scans per session; 
        $\mathrm{N_T}$: average non-repeated time points per subject; 
        Age is reported as mean ± standard deviation. 
        Disease categories include cognitively normal (CN), mild cognitive impairment (MCI), Alzheimer’s disease (AD), and other diagnoses. 
        Field strength refers to scanner magnetic field strength (1.5T, 3T). 
        Manufacturer categories include Siemens Healthineers (SH), General Electric (GE), Philips Healthcare (PH). If the summed counts across classes, field strengths, or manufacturers are lower than $\mathrm{N_I}$, this is due to missing metadata. Scans lacking the corresponding labels were excluded only from the subsequent evaluations.}
    \end{tablenotes}
\end{threeparttable}
\end{table*}
}

Five publicly available datasets of T1w MRI scans were used in this study:

\begin{itemize}
    \item The \textbf{ADNI}\footnote{\scriptsize Data used in the preparation of this article were obtained from the Alzheimer’s Disease Neuroimaging Initiative (ADNI) database (\url{http://adni.loni.usc.edu}). The ADNI was launched in 2003 as a public-private partnership, led by Principal Investigator Michael W. Weiner, MD. The primary goal of ADNI has been to test whether serial magnetic resonance imaging (MRI), positron emission tomography (PET), other biological markers, and clinical and neuropsychological assessment can be combined to measure the progression of mild cognitive impairment (MCI) and early Alzheimer’s disease (AD). For up-to-date information, see~\url{http://www.adni-info.org}.}    
    dataset is a comprehensive resource comprising clinical, imaging, genetic, cognitive, and biomarker data from individuals with varying cognitive statuses, including CN, mild cognitive impairment (MCI), and AD \citep{jack_alzheimers_2008}. For this study, we used \textgreater\num[round-mode=places,round-precision=0]{20971} images from 2422 subjects. Data were acquired longitudinally using scanners from multiple manufacturers, \reviiadd{including Siemens, General Electric, and Philips}, at field strengths of 1.5T and 3T (see Table~\ref{tab:dem_des}). This dataset contains repeated MRI scans acquired within the same imaging session, providing multiple consecutive acquisitions of the same subject under identical conditions (denoted as $\mathrm{N_R}$ in Table~\ref{tab:dem_des}). These repetitions help assess within-session variability and improve the reliability of longitudinal analyses. 
    
    \item The \textbf{OASIS3} (Open Access Series of Imaging Studies 3) is a longitudinal, multimodal neuroimaging dataset that integrates clinical, cognitive, and biomarker data to study normal aging and AD \citep{oasis3}. As stated on Table~\ref{tab:dem_des}, the dataset includes 1316 participants with 2681 MRI sessions acquired on Siemens scanners at 1.5T and 3T.
    
    \item The \textbf{MIRIAD} (Minimal Interval Resonance Imaging in Alzheimer's Disease) dataset provides longitudinal T1w MRI scans from 46 patients with mild-to-moderate AD and 23 healthy elderly controls \citep{malone_miriadpublic_2013}. Scans were acquired at short intervals (2 weeks to 2 years) using \reviiadd{General Electric} scanners at 1.5T (see Table~\ref{tab:dem_des}). This dataset contains repeated MRI scans for the same session.
    
    \item The \textbf{AIBL}\footnote{\scriptsize Data was collected by the AIBL study group. AIBL study methodology has been reported previously (Ellis et al. 2009).}
    dataset combines MRI and PET imaging with clinical, cognitive, genetic, and lifestyle measures \citep{ellis_australian_2009}. For this work, we used 1296 T1w scans from 691 participants either at risk for or diagnosed with AD. Data were acquired on Siemens scanners at 1.5T and 3T (see Table~\ref{tab:dem_des}).
    
    \item The \textbf{SLIM} (Southwest University Longitudinal Imaging Multimodal) dataset includes multimodal MRI data and behavioral assessments from healthy young adults of around 21 years old \citep{liu_longitudinal_2017}. The longitudinal study spans 3.5 years, with 3 time points to study brain–behavior correlations and test–retest reliability. It contains 1037 T1w scans from 588 subjects, acquired on Siemens 3T scanners, as shown in Table~\ref{tab:dem_des}.
\end{itemize}

To quantitatively evaluate the differences among datasets, we computed the Fréchet AutoEncoder Distance (FAED) between each pair of datasets using features extracted from a pretrained 3D medical image autoencoder 
\citep{buzuti_frechet_2023}. This metric, inspired by the widely used Fréchet Inception Distance in the GAN literature \citep{heusel_gans_2018}, measures the distance between the feature distributions of two datasets and is sensitive to differences in image quality and content \citep{heusel_gans_2018}.

\begin{figure}[h]
  \centering
  \includegraphics[width=\columnwidth]{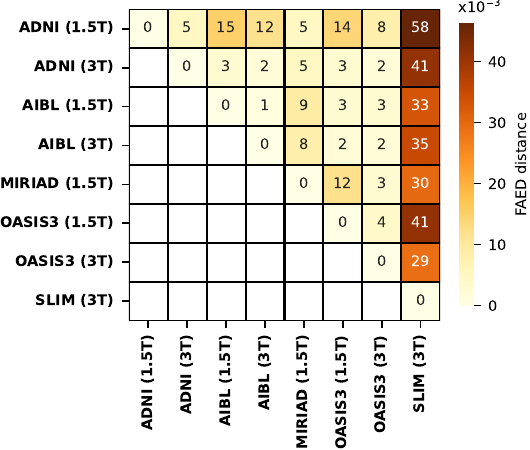}
  \caption{Pairwise FAED between MRI datasets stratified by MRI field strength (1.5T and 3.0T). Lower values indicate greater similarity in feature distributions, while higher values reflect larger domain shifts. White cells correspond to uncomputed comparisons. The comparisons were done with the fully-preprocessed brain images.}
  \label{fig:FAED}
\end{figure}

For each dataset, 200 samples were randomly selected. Then, the FAED between datasets was computed in the feature space of a pretrained 3D ResNet-10 from MONAI \citep{hara_can_2018, He2016, chen_med3d_2019, cardoso_monai_2022}. The results are summarized in Figure \ref{fig:FAED}. We observed relatively low distances among the ADNI, AIBL, MIRIAD, and OASIS3 datasets, meaning these four datasets are largely comparable. In contrast, the SLIM dataset showed higher distances with the other datasets. We attribute these higher distances to the noticeably younger population for the SLIM dataset compared to the other four datasets. 

\subsection{Imaging prepossessing}

We followed the preprocessing protocol described in \citet{fu2025synthesizing}. Specifically, we used skull-stripped and bias field–corrected T1w images processed \reviiadd{according to the FreeSurfer process flow\footnote{\scriptsize \url{https://surfer.nmr.mgh.harvard.edu/fswiki/ReconAllDevTable}}} (i.e., \path{norm}, 1 \path{mm} isotropic). We affinely registered each image to the FreeSurfer Talairach space. We then rescaled image intensities to [0,1] and cropped them to [160,~176,~208] \reviiadd{, starting from [48,~38,~10]}. We determined this crop size to preserve most foreground information based on 200 randomly selected scans.

We used FreeSurfer-processed brain segmentation masks (i.e., \path{aseg+aparc}) for training and evaluation. For the ADNI and OASIS3 datasets, we obtained segmentations from FreeSurfer v5.3.0 in TheHiveDB platform \citep{muehlboeck_thehivedb_2014}. We segmented the AIBL, MIRIAD, and SLIM datasets using FastSurfer\footnote{\scriptsize FastSurfer is a deep learning–based reimplementation of FreeSurfer.} v2.4.2 \citep{henschel_fastsurfer_2020}. These masks delineate cortical and subcortical regions, enabling both global and region-specific analyses. In total, we included 103 cortical and subcortical regions in this study. 

\reviiadd{As part of quality control, images with low mutual information with respect to the FreeSurfer Talairach template were visually inspected, and scans showing evident preprocessing failures were excluded \citep{iglesiasReadytouseMachineLearning2023}. The data reported in Table~\ref{tab:dem_des} correspond to the dataset after this screening step.}

A dataset of brain regions was derived from all original brain datasets. 
We defined a common bounding box for each of the 103 brain regions using 500 randomly selected brains. We then segmented and cropped all regions based on these bounding boxes and resampled them to a universal size of [64, 64, 64] using trilinear interpolation. As an example, we cropped the left hippocampus using coordinates [57,~31,~0] to [153,~175,~192], yielding an initial size of [96, 144, 192], which we subsequently resampled to [64,~64,~64].

\subsection{Dataset setup}

We used the ADNI and OASIS3 datasets for training, \reviadd{validation} and testing, while reserving the AIBL, MIRIAD, and SLIM datasets  \reviiadd{exclusively for external} testing. \reviiadd{The ADNI and OASIS3 datasets were split at the subject level,} allocating approximately 80\% of subjects for training, \reviadd{10\% for validation} and \reviadd{10}\revidel{20}\% for testing. Each partition was stratified to maintain the same proportion of dementia patients and healthy subjects. \reviadd{For subjects with longitudinal scans, all MRI examinations from the same individual were assigned the label corresponding to the most recent available clinical assessment. Notice that during training, we do not use the class, only the subject identity.}

Two different datasets were considered: one comprising whole-brain scans and another containing segmented brain regions. These were used to train and evaluate the two separate CBIR frameworks.

\begin{figure*}[!htbp]
  \centering
  \includegraphics{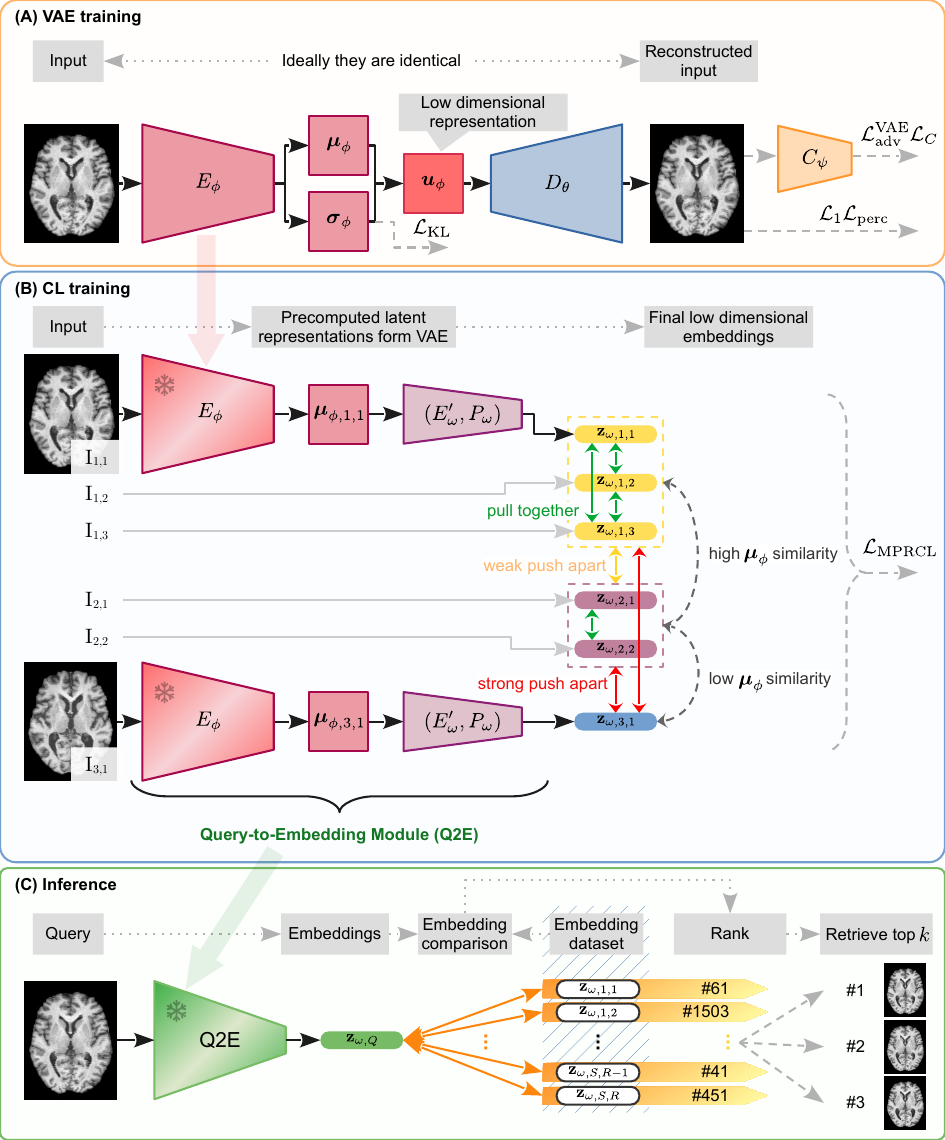}
  \caption{
Overview of the NeuroCBIR framework. 
(A) VAE training: the encoder ($E_\phi$) maps input images to mean ($\boldsymbol{\mu}_\phi$) and variance ($\boldsymbol{\sigma}_\phi$), which define the latent variable ($\boldsymbol{u}_\phi$). The decoder ($D_\theta$) reconstructs the image, optimized using an $L_1$ loss, latent regularization ($L_{\mathrm{KL}}$), perceptual loss, and an adversarial loss with critic ($C_\psi$). 
(B) CL training: the latent representations, i.e. the means $\boldsymbol{\mu}_\phi$ from the pretrained VAE encoder, are passed through an encoder–projector module $(E'_\omega, P_\omega)$, yielding compact embeddings ($\boldsymbol{z}_{\omega}$). Similar (intra-subject) embeddings are pulled together, while dissimilar (inter-subject) embeddings are pushed apart by the criteria specified for the Multi-Positive Ranking Contrastive Loss ($L_{MPRCL}$). We denote by $\mathrm{I}_{i,j}$ the input image corresponding to subject $S$ and index $R$.
(C) Inference: the frozen Q2E module maps a query image to its embedding, which is compared against the precomputed database \{$\mathbf{z}_{\omega,1,1}$, $\mathbf{z}_{\omega,1,2}$, ..., $\mathbf{z}_{\omega,S,R-1}$, $\mathbf{z}_{\omega,S,R}$\} to rank and retrieve the top-$k$ most similar images.}

  \label{fig:flowchart}
\end{figure*}

\subsection{Framework of NeuroCBIR}

Figure~\ref{fig:flowchart} presents a schematic overview of the NeuroCBIR training and inference workflow. In the first training stage, NeuroCBIR extracts latent representations from input images using a VAE. Then, these precomputed representations are passed to an encoder–projector model with contrastive learning to pull together intra-subject (similar) images and push apart inter-subject (dissimilar) ones. The resulting compact embeddings enable NeuroCBIR to efficiently perform similarity comparisons and retrieve the top-$k$ most similar images during inference.

\subsubsection{VAE-GAN}

We used a VAE to learn latent representations because of its inference efficiency \citep{rombach_high-resolution_2021, pinaya_brain_2022}. Specifically, we followed the VAE-GAN framework from \citet{rombach_high-resolution_2021, pinaya_brain_2022} and trained two separate models for whole-brain and brain-regional data, respectively.

We used the \path{AutoencoderKL} model from the MONAI library \citep{cardoso_monai_2022, rombach_high-resolution_2021, pinaya_brain_2022}. It uses an encoder–bottleneck–decoder structure, where both encoder $E$ and decoder $D$ implement four residual convolutional layers with 64, 128, 128, and 128 filters, using group normalization. The discriminator $C$ implements three convolutional layers of 32 filters each, using instance normalization. 

Each encoder block reduces the spatial resolution of the input volume through strided convolutions, progressively compressing the feature maps into a latent representation with 8 channels. The resulting latent tensor size thus depends on the input field of view: for the whole-brain input, the downsampling yields a latent shape of $[8,\,20,\,22,\,26]$, while for brain-region inputs, it results in $[8,\,8,\,8,\,8]$.

The encoder, parameterized by $\phi$, maps an input volume $\mathbf{x}$, drawn from the empirical data distribution $p_{\text{data}}$ with training samples 
$\mathrm{\{I_n\}_{n=1}^{N_I}}$, to an approximate posterior 
distribution over latent variables:
\begin{equation}
q_\phi(\mathbf{u}\mid \mathbf{x}) = \mathcal{N}\!\left(\boldsymbol{\mu}_\phi(\mathbf{x}),\,\operatorname{diag}\boldsymbol{\sigma}^2_\phi(\mathbf{x})\right),
\end{equation}
while the prior is given by $p(\mathbf{u})=\mathcal{N}(\mathbf{0},\mathbf{I})$.
A latent sample $\mathbf{u}$ is drawn using the reparameterization
$\mathbf{u}
=\boldsymbol{\mu}_\phi(\mathbf{x})
+\boldsymbol{\sigma}_\phi(\mathbf{x})\odot\boldsymbol{\varepsilon}$,
with $\boldsymbol{\varepsilon}\!\sim\!\mathcal{N}(\mathbf{0},\mathbf{I})$. The decoder $D_\theta$ reconstructs the input as $\hat{\mathbf{x}} = D_\theta(\mathbf{u})$. A PatchDiscriminator $C_\psi$ provides an adversarial signal that encourages realistic reconstructions. 

\paragraph{VAE-GAN training} To stabilize training and fit a larger effective batch size within GPU memory limits, we used a base batch size of 1 with 8 gradient accumulation steps. We trained both the autoencoder $(\phi,\theta)$ and the discriminator $\psi$ using the Adam optimizer with a learning rate of $1\times10^{-4}$. We employed mixed--precision training to improve computational efficiency. The VAE was trained for 7 full epochs of the training partition.

We optimized the autoencoder parameters $(\theta,\phi)$ with a weighted combination of four terms: (i) Reconstruction loss: for each input $\mathbf{x}$ defined on a voxel grid $\Omega$, we minimize the mean absolute error $\mathcal{L}_{1}(\mathbf{x})=\frac{1}{|\Omega|}\, \mathbb{E}_{\mathbf{u}\sim q_\phi(\mathbf{u}\mid\mathbf{x})} \big[\lVert \mathbf{x}-D_\theta(\mathbf{u})\rVert_{1}\big]$; (ii) Kullback--Leibler divergence: the latent posterior is regularized toward the unit Gaussian prior, $\mathcal{L}_{\mathrm{KL}}(\mathbf{x})=D_{\mathrm{KL}}\!\big(q_\phi(\mathbf{u}\mid\mathbf{x}) \,\Vert\, p(\mathbf{u})\big)$; (iii) Perceptual loss: $\mathcal{L}_{\mathrm{perc}}$ is computed as the feature--space $\ell_1$ distance between $\mathbf{x}$ and $\hat{\mathbf{x}}$ using activations from a pretrained medical imaging network; and (iv) Adversarial loss: based on the least--squares GAN formulation, $\mathcal{L}_{\mathrm{adv}}^{\mathrm{VAE}} = \mathbb{E}_{\mathbf{x}\sim p_{\text{data}}}\!\left[(C_\psi(D_\theta(\mathbf{u}))-1)^2\right]$, 
with $\mathbf{u}\!\sim\!q_\phi(\mathbf{u}\mid\mathbf{x})$.
The total generator (autoencoder) objective is therefore:
\begin{equation}
\begin{split}
\mathcal{L}_{\mathrm{VAE}} &= \lambda_{1}\mathcal{L}_{1} 
+ \lambda_{\mathrm{perc}}\mathcal{L}_{\mathrm{perc}} \\
&\quad + \lambda_{\mathrm{KL}}\mathcal{L}_{\mathrm{KL}} 
+ \lambda_{\mathrm{adv}}\mathcal{L}_{\mathrm{adv}}^{\mathrm{VAE}}.
\end{split}
\end{equation}

The discriminator is trained with the least-squares objective:
\begin{equation}
\begin{aligned}
\mathcal{L}_{C}
&=\tfrac{1}{2}\,\mathbb{E}_{\mathbf{x}\sim p_{\text{data}}}
  \!\left[(C_\psi(\mathbf{x})-1)^2\right] \\
&\quad + \tfrac{1}{2}\,\mathbb{E}_{\mathbf{x}\sim p_{\text{data}},\, 
\mathbf{u}\sim q_\phi(\mathbf{u}\mid\mathbf{x})}
  \!\left[C_\psi(D_\theta(\mathbf{u}))^2\right],
\end{aligned}
\end{equation}
with loss weights $\lambda_{1}=1.0$,
$\lambda_{\mathrm{perc}}=0.1$,
$\lambda_{\mathrm{adv}}=0.1$, and $\lambda_{\mathrm{KL}}=1\times 10^{-7}$ \citep{rombach_high-resolution_2021, pinaya_brain_2022}.

Both the autoencoder $(\phi,\theta)$ and the discriminator $\psi$ were trained iteratively in an alternating manner at each batch. In each iteration, the autoencoder parameters $(\phi,\theta)$ were first updated using $\mathcal{L}_{\mathrm{VAE}}$, while keeping the discriminator $\psi$ fixed. Then, the discriminator $\psi$ was updated to distinguish real from reconstructed samples using $\mathcal{L}_{C}$.

\subsubsection{Contrastive Learning}

After training the VAE--GAN, we extracted the latent mean representations $\boldsymbol{\mu}_\phi$ for all input images. These representations were subsequently used to train an encoder--projector module $(E'_\omega, P_\omega)$ with learnable parameters $\omega$ using a contrastive learning objective. Here, $E'_\omega$ is a convolutional encoder operating on the latent representations, while $P_\omega$ is a fully connected projection head. Using $\boldsymbol{\mu}_\phi$ instead of raw images enables larger batch sizes; in this work, a batch size of 128 samples is used for both whole-brain and region-based approaches.

For each sample $i$, the latent vector $\boldsymbol{\mu}_{\phi,i}(\mathbf{x}_i)$ is projected into a compact embedding space as:
\begin{equation}
\mathbf{z}_{\omega,i} = P_\omega\big(E'_\omega(\underbrace{\boldsymbol{\mu}_{\phi,i}(\mathbf{x}_i)}_{\text{precomputed}}\big) \in \mathbb{R}^\mathrm{N_f},
\end{equation}
where $\mathrm{N_f}$ denotes the dimensionality of the learned embedding space. \reviadd{The embedding dimensionality was empirically optimized through a hyperparameter tuning study by evaluating $\mathrm{N_f} \in \{16, 32, 64, 128, 256\}$ (see Section~\ref{subsec:results_hyp_op}). } The embeddings are $\ell_2$-normalized to unit length, i.e., $\|\mathbf{z}_{\omega,i}\|_2 = 1$.

Given a batch of $B$ samples, we compute the pairwise cosine similarity matrix:
\begin{equation}
s_{ij} = \mathbf{z}_{\omega,i}^\top \mathbf{z}_{\omega,j}, \quad i,j = 1,\ldots,B.
\end{equation}

\begin{figure}[!htbp]
  \centering
  \includegraphics[width=\columnwidth]{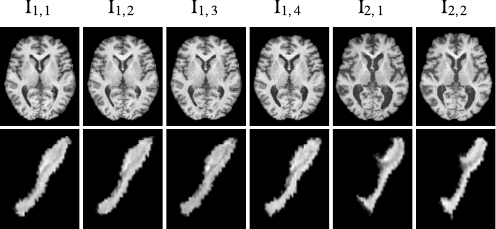}
  \caption{\small Example of input and target images for whole-brain and brain-region data. We denote by $\mathrm{I}_{S,R}$ the input image corresponding to subject $S$ and index $R$ (i.e., time point). $\mathrm{I}_{1,1}$--$\mathrm{I}_{1,4}$ show four images for the subject 1, and $\mathrm{I}_{2,1}$--$\mathrm{I}_{2,2}$ show two images for the subject 2. The brain-region images illustrate the hippocampus.}
  \label{fig:example_brain_region_sample}
\end{figure}

Each sample $i$ is associated with a categorical label $y_i$, which defines \emph{hard positives} as samples sharing the same label ($y_i = y_j$) and \emph{negatives} as samples with different labels. For whole-brain images, the label corresponds to subject identity, whereas for brain-region images it corresponds to the combination of subject and brain region. Figure~\ref{fig:example_brain_region_sample} shows examples of whole-brain images and brain-region images from two different subjects.

\begin{figure}[!htbp]
  \centering
  \includegraphics[width=\columnwidth]{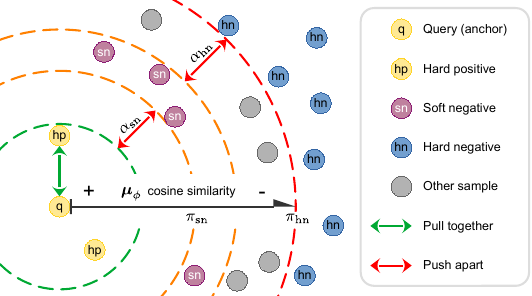}
  \caption{Illustration of the proposed Multi-Positive Ranking Contrastive Loss (MPRCL). 
For a given anchor query, hard positives are defined by shared identity labels, while soft and hard negatives are selected using percentile thresholds applied to the VAE latent similarity. Soft negatives correspond to samples whose VAE similarity exceeds the upper percentile $\pi_{\mathrm{soft}}$, whereas negatives fall below the lower percentile $\pi_{\mathrm{hn}}$. 
In the embedding space, the loss enforces margin-based ranking constraints that encourage hard positives to be closest to the anchor, soft negatives to occupy an intermediate similarity range, and negatives to be pushed further away. The margins are adaptively scaled by the anchor-specific similarity range, using scaling factors $\alpha_{\mathrm{id}}$ for identity-based constraints and $\alpha_{\mathrm{soft}}$ for soft-negative versus negative constraints.}
  \label{fig:loss_function}
\end{figure}

\paragraph{Multi-Positive Ranking Contrastive Loss}
We propose a Multi-Positive Ranking Contrastive Loss (MPRCL) that explicitly models three types of relationships for each anchor sample: (i) hard positives, corresponding to samples from the same subject; (ii) soft negatives, defined as different-subject samples that are close in latent space; and (iii) hard negatives, defined as different-subject samples that are far in latent space (see Figure~\ref{fig:loss_function}). MPRCL enforces margin-based ranking constraints that induce a structured ordering of similarities in the embedding space \citep{tang_ranking-based_2023}.

\paragraph{Soft-negative and hard-negative mining}
Soft and hard negatives are identified at each training iteration using the latent representations produced by the VAE. A VAE-based cosine similarity matrix is computed as:
\begin{equation}
\tilde{s}_{ij} =
\frac{\boldsymbol{\mu}_{\phi,i}^\top \boldsymbol{\mu}_{\phi,j}}
{\|\boldsymbol{\mu}_{\phi,i}\|_2 \, \|\boldsymbol{\mu}_{\phi,j}\|_2},
\end{equation}
which is treated as a fixed, non-trainable similarity signal within the batch. For each anchor $i$, soft negatives are defined as label-negative samples whose VAE similarity exceeds an upper percentile threshold $\pi_{\mathrm{sn}}$, while hard negatives are defined as label-negative samples whose VAE similarity falls below a lower percentile threshold $\pi_{\mathrm{hn}}$. Hard-positive samples are excluded from both mining procedures. Percentiles are computed independently for each anchor, allowing the mining process to adapt to the local similarity distribution within each batch.

\paragraph{Margin scaling}
To ensure robustness across batches with varying similarity distributions, margin values are dynamically scaled for each anchor at every batch. Let
\begin{equation}
\Delta_i = \max_{j \neq i} s_{ij} - \min_{j \neq i} s_{ij}
\end{equation}
denote the similarity range for anchor $i$, excluding self-similarity. The soft-negative and hard-negative margins are defined as:
\begin{equation}
m_i^{\mathrm{sn}} = \alpha_{\mathrm{sn}} \Delta_i,
\qquad
m_i^{\mathrm{hn}} = \alpha_{\mathrm{hn}} \Delta_i,
\end{equation}
where $\alpha_{\mathrm{sn}}$ and $\alpha_{\mathrm{hn}}$ are fixed hyperparameters. Hard positives are excluded from the margin computation.

\paragraph{Hard-positive attraction loss}
For each anchor $i$, let $s_i^{\mathrm{hp,min}}$ denote the minimum similarity to its hard positives. The hard-positive attraction loss is defined as:
\begin{equation}
\mathcal{L}_{\mathrm{hp}}^{(i)} =
\left[1 - s_i^{\mathrm{hp,min}}\right]_+,
\end{equation}
where $[\cdot]_+ = \max(0, \cdot)$. Each batch is constructed to contain at least 32 anchors with a minimum of three hard-positive samples per anchor.

\paragraph{Soft-negative ranking loss}
To further separate hard positives from soft negatives, we define:
\begin{equation}
\mathcal{L}_{\mathrm{sn}}^{(i)} =
\left[m_i^{\mathrm{sn}} + s_i^{\mathrm{sn,max}} - s_i^{\mathrm{hp,min}}\right]_+,
\end{equation}
where $s_i^{\mathrm{sn,max}}$ denotes the maximum similarity between anchor $i$ and its soft negatives.

\paragraph{Hard-negative ranking loss}
To further separate soft negatives from hard negatives, we impose the constraint:
\begin{equation}
\mathcal{L}_{\mathrm{hn}}^{(i)} =
\left[m_i^{\mathrm{hn}} + s_i^{\mathrm{hn,max}} - s_i^{\mathrm{sn,min}}\right]_+,
\end{equation}
where $s_i^{\mathrm{sn,min}}$ denotes the minimum similarity among the soft negatives for anchor $i$, and $s_i^{\mathrm{hn,max}}$ denotes the maximum similarity to its hard negatives.

\paragraph{Final objective}
The overall loss is obtained by averaging over all anchors and combining the three components:
\begin{equation}
\mathcal{L}_{\mathrm{MPRCL}} =
\frac{1}{B}\sum_{i=1}^{B} \mathcal{L}_{\mathrm{hp}}^{(i)}
+
\lambda_{\mathrm{sn}} \frac{1}{B}\sum_{i=1}^{B} \mathcal{L}_{\mathrm{sn}}^{(i)}
+
\lambda_{\mathrm{hn}} \frac{1}{B}\sum_{i=1}^{B} \mathcal{L}_{\mathrm{hn}}^{(i)},
\end{equation}
where $\lambda_{\mathrm{sn}}$ and $\lambda_{\mathrm{hn}}$ control the relative contribution of the ranking terms.

In the region-based approach, the loss $\mathcal{L}_{\mathrm{MPRCL}}$ is computed independently per brain region. At each training iteration, 10 regions are randomly selected from the total set of 103 regions within the batch. The corresponding regional losses are computed and then averaged to obtain the final loss value.

This formulation encourages a structured embedding space in which longitudinal scans of the same subject are tightly clustered, semantically similar samples occupy intermediate distances, and dissimilar samples are pushed apart. In the experiments in Section~\ref{sec:results}, the hyperparameters are optimized via grid search to obtain their optimal configuration.

Optimization was performed using the AdamW optimizer, with an initial learning rate of $1\times10^{-4}$. Training was terminated once the training loss exhibited no further noticeable improvement.


\subsubsection{Inference}

Once the VAE and the subsequent encoder–projector are trained, the entire block---represented in Figure~\ref{fig:flowchart} as the Query-to-Embedding (Q2E) module---is used to generate embeddings for the dataset. Specifically, embeddings are extracted from all images, creating a reference embedding database.

Given a new query image, the Q2E module produces its corresponding embedding, which is compared against the reference embeddings using pairwise cosine similarity. The similarity scores are then ranked, and the top-$k$ most similar images are retrieved.

\subsection{Evaluation metrics}  
We assess the performance of our model using complementary metrics that capture both retrieval accuracy and the quality of the learned embedding space.

\subsubsection{Re-identification metrics}
We use two complementary metrics to assess the retrieval quality. First, the mean average precision at $k$ (mAP@$k$) measures the average proportion of relevant images among the top-$k$ results for each query, reflecting how well the system ranks similar cases. In turn, the mean success at $k$ (mS@$k$) indicates the fraction of queries for which at least one relevant image appears in the top-$k$, showing how often the system retrieves at least one correct match. The mathematical formulations of the metrics used are \citep{giner_intrinsic_2024}:  
\begin{equation}
\text{Precision@}k_i = \text{Prec@}k_i = \frac{|{\rm rel}_i^{(k)}|}{k}
\end{equation}

\begin{equation}
\text{mAP}@k = \frac{1}{\mathrm{N_{queries}}} \sum_{i=1}^{\mathrm{N_{queries}}} \text{Prec@}k_i
\end{equation}

\begin{equation}
\begin{aligned}
\text{mS@}k &= \frac{1}{\mathrm{N_{queries}}} \sum_{i=1}^{\mathrm{N_{queries}}} s_i, \\
s_i &=
\begin{cases}
1, & \text{if query $i$ has a hit in top-$k$}, \\
0, & \text{otherwise.}
\end{cases}
\end{aligned}
\end{equation}
where $|{\rm rel}_i^{(k)}|$ denotes the number of relevant images among the top-$k$ retrieved for query $i$, and $\mathrm{N_{queries}}$ is the total number of queries. We define a retrieval as a “hit” when the retrieved image belongs to the same subject as the query image.

We evaluated both metrics on several query subsets: all images, training and test partitions, and each dataset separately (ADNI, OASIS3, AIBL, MIRIAD, and SLIM). For every evaluation, each image in the selected subset served once as a query, while the reference database for retrieval always contained the entire image collection. Cosine similarity between image embeddings determined the ranking of retrieved images. 

When computing metrics for a given $k$ (e.g., $k=5$), queries corresponding to subjects with fewer than $k$ relevant images in the database were excluded from the evaluation. This ensures that all metrics are computed on queries for which a top-$k$ retrieval is meaningful.

\subsubsection{Embedding-space similarity}
In addition to retrieval-based metrics, we assess the quality of the learned embedding space using Spearman's rank correlation coefficient. This metric quantifies the monotonic relationship between pairwise similarities in the embedding space and reference similarity measures, without assuming linearity.

For a set of 100 image pairs $(i,j)$ per dataset, we compute the cosine similarity between embeddings, $s_{ij}$, and compare it to the multi-scale structural similarity index (MS-SSIM), $s_{\mathrm{MS-SSIM},ij}$. Spearman's correlation is then computed as:
\begin{equation}
\rho_{\mathrm{MS-SSIM}} = 1 - \frac{6 \sum_{(i,j)} \Big(\text{rank}(s_{ij}) - \text{rank}(s_{\mathrm{MS-SSIM},ij})\Big)^2}{(n^2 - 1)},
\end{equation} 
where $n$ is the number of evaluated image pairs, i.e., the total number of images in the dataset excluding those from the same subject as the query.

Because samples are ranked such that lower ranks correspond to higher similarity, a strong agreement between embedding similarity and MS-SSIM will result in a negative correlation. Specifically, $\rho \approx -1$ indicates that the embedding similarities closely match the perceptual similarity rankings, $\rho \approx 0$ indicates no correlation, and $\rho \approx 1$ would indicate an inverse relationship.

\subsection{Interpretability}
Region-based retrieval is inherently interpretable because similarity can be directly linked to specific, anatomically defined brain regions. In contrast, whole-brain embeddings are high-dimensional and spatially distributed, making it difficult to identify which regions drive similarity-based retrieval. To better understand the regional contributions underlying whole-brain similarity, we performed a correlation analysis between whole-brain and region-specific similarities.

A set of 100 whole-brain query scans was randomly selected. For each query, we retrieved the top-50, top-200, and top-1000 most similar cases based on the whole-brain embedding, excluding scans from the same subject. For these retrieved cases, cosine similarity scores were also computed independently for each anatomical region using the region-based retrieval pipeline.

To quantify each region's contribution to whole-brain similarity, we computed the Spearman rank correlation between whole-brain similarity scores and region-specific similarity scores across the retrieved samples. Intuitively, a high correlation for a given region indicates that the region alone can largely explain the similarity ranking produced by the whole-brain embedding. Correlations were computed separately for each top-$k$ setting.

Finally, to summarize variability in regional contributions, we computed the coefficient of variation (CV), defined as the ratio of the standard deviation to the mean of the Spearman correlations. CV was evaluated both across regions for each query and across queries for each region. We interpret the CV values as follows:

\begin{itemize}
    \item Low CV across regions for a query indicates that all regions contribute consistently to retrieving the top-$k$ matches for that query.
    \item High CV across regions for a query indicates that individual regions vary in their influence on whole-brain similarity for that query.
    \item Low CV across queries for a region indicates that this region consistently contributes to the top-$k$ retrieval across different queries, i.e., the retrieval process relies more heavily on this region in general.
\end{itemize}

\subsection{Bias evaluation}  

Each dataset was acquired under different study protocols and included different populations. Moreover, within a dataset, MRI acquisition parameters (e.g., scanner manufacturer, field strength) also varied. Such differences may introduce systematic biases in retrieval performance.  

To investigate these effects, the NeuroCBIR system was evaluated using query subsets defined by scanner field strength (1.5T or 3T) and by manufacturer (GE, Philips, or Siemens). In these analyses, only images from a given subset served as queries, while retrieval was performed against the full database. For example, when evaluating 1.5T queries, all 1.5T images were used as queries, but the reference database included all images, keeping the pool of candidate images constant across subsets.

For each query, the Prec@5 was computed as the proportion of relevant images among the top-5 retrieved results. Collecting the Prec@5 values across all queries yields a distribution of retrieval performance for each subset (e.g., 1.5T, 3T, GE, Philips, and Siemens). Since the distributions of Prec@5 along queries did not follow normality, the non-parametric test of {Mann--Whitney~U test} was used to assess statistically significant differences between subsets. Additionally, {Cliff’s $\delta$} was calculated to quantify the effect size, providing a measure of the magnitude of differences independent of sample size. According to conventional thresholds, $|\delta|<0.15$ is considered negligible, $0.15 \leq |\delta| < 0.33$ is small, $0.33 \leq |\delta| < 0.47$ is medium, and $|\delta| \geq 0.474$ is large \citep{meissel_using_nodate}.

\subsection{Comparison with pretrained models}

To benchmark the performance of our proposed NeuroCBIR system, we evaluated several commonly used 3D feature extractors trained on medical imaging cohorts. The tested models included standard 3D ResNet architectures (ResNet-10, ResNet-50, ResNet-101) from MONAI \citep{hara_can_2018, He2016, chen_med3d_2019, cardoso_monai_2022}, a 3D-CNN pre-trained model for brain T1-weighted MRI for AD diagnosis (CNN-AD) \citep{turrisi_deep_2024}, and a large-scale medical vision model based on the transformer Swin UNETR (VoComni-L) \citep{hatamizadeh_swin_2022,wu_voco_2024, wu_large-scale_2024, cardoso_monai_2022}. For all ResNet models and VoComni-L, we extracted features from the deepest encoder layers, followed by global average pooling to obtain fixed-size embeddings. Additionally, we evaluated the performance of our VAE-based embeddings projecting with UMAP \citep{mcinnes_umap_2020}. 

\reviadd{Additionally, two 3D brain foundational models were added to the comparison: BrainIAC \citep{takGeneralizableFoundationModel2026} and FOMO2JOMO \citep{munkBrainMRIFoundation2026}. BrainIAC was trained on \num[round-mode=places,round-precision=0]{48965} brain MRIs with different contrasts (T1w, T2w, T2-FLAIR, contrast-enhanced T1w) across a broad spectrum of tasks. This model 
outperformed localized supervised training and other pretrained models, particularly in low-data, few-shot settings and in high-difficulty prediction tasks, allowing for application in scenarios otherwise infeasible. 
The method leverages self-supervised foundation models pretrained using SimCLR and masked autoencoders  on 3D brain MRI. 
For the present work, all images were adapted to fit with their preprocessing steps \citep{takGeneralizableFoundationModel2026}. In turn, FOMO2JOMO, was the winner of the MICCAI 2025 challenge FOMO25 
This method uses a similar loss function to ours for training their model, i.e, they try to pull together images from the same subject while pushing apart those from different modalities. The method pretrained a 3D CNN multi-modal UNet-style VAE with 18.9M parameters. 
This model uses patches of size [$64, 64, 64$] rather than the entire image \citep{munkBrainMRIFoundation2026}. Thus, we used the central cropped brain volume to extract the mean values of the subject-level latent representations in the VAE bottleneck. Every mean value has a raw dimension of [$256, 4, 4, 4$] and were global averaged pooled to 256 features.
}

The extracted embeddings from each model were used to perform CBIR on the testing data, reporting the compression rate, the number for features, mAP@5, mS@5 and $\rho_{\mathrm{MS-SSIM}}$.

\subsection{Computational complexity}  

The execution time of the NeuroCBIR system is critical for assessing its feasibility in real-time applications. In clinical practice, the dataset of embeddings is precomputed in advance. Therefore, during inference, the main factors that determine efficiency are loading the Q2E module, processing the query to generate its embedding, and computing the pairwise cosine similarity with the precomputed embedding dataset, followed by sorting and retrieving the top matches. 

All execution times reported in the current study were obtained on a system with the following characteristics: PyTorch was configured to use 1 and 4 CPU threads. The platform was Linux 4.18.0 running on an x86\_64 machine with an Intel\textsuperscript{\textregistered} Xeon\textsuperscript{\textregistered} Gold 6226R CPU~@~2.90~GHz.

\subsection{Application-specific experimental setups}
To further assess the clinical relevance of the learned embeddings beyond retrieval performance, we conducted application-specific evaluations. These analyses test whether the NeuroCBIR framework captures meaningful biological and disease-related patterns in a fully zero-shot setting. 

Zero-shot evaluation refers to applying the learned NeuroCBIR embeddings to new tasks, such as age estimation or disease risk inference, without any task-specific training or fine-tuning, allowing us to assess whether the embeddings capture clinically meaningful information in a fully generalizable and unbiased manner \citep{guo_interpretable_2025, bazine_convolutional_2024}.

\subsubsection{Zero-shot similarity-based age inference}
\label{sec:zero_shot_age}

We evaluated whether the NeuroCBIR framework captures age-related morphological patterns without any age or disease supervision. To isolate normative aging effects, the retrieval database was restricted to CN subjects. For scans with repeated acquisitions, only a scan per time point was retained to reduce redundancy.

To mitigate the impact of sparse age coverage below 50 years, the retrieval database included only subjects older than 50. The query set was further constrained to subjects whose ages fell between the 5th and 95th percentiles. Two query sets were constructed (CN and AD subjects), while a shared CN reference pool was used for retrieval in both cases.

Given a query MRI scan, NeuroCBIR retrieved the top-$k$ most similar CN scans (excluding scans from the same subject) using the learned embedding space. Age prediction was performed by aggregating the ages of the retrieved neighbors, without training an explicit regression model.

Let $a_i$ denote the chronological age of the $i$-th retrieved neighbor, and let $p(a_i)$ denote the empirical age distribution in the CN reference set. To compensate for non-uniform age sampling, each neighbor was assigned an inverse-frequency weight proportional to $1 / p(a_i)$. The predicted age $\hat{a}$ was then computed as:

\begin{equation}
\hat{a} = \frac{\sum_{i=1}^{k} a_i \, p(a_i)^{-1}}{\sum_{i=1}^{k} p(a_i)^{-1}}.
\end{equation}

Age inference performance was quantified using the mean absolute error (MAE) between predicted and chronological age. Pearson’s correlation coefficient ($r$) and the regression slope were also computed to assess the strength and calibration of the predictions. This evaluation was applied independently for each anatomical region and for the whole-brain embedding.

\subsubsection{Zero-shot similarity-based disease risk estimation}

We also applied a zero-shot neighbor-based approach to estimate disease risk, without supervised disease classification or fine-tuning. The retrieval set comprised all available MRI scans, excluding scans from the same subject as the query. For scans with repeated acquisitions, only a scan per time point was retained to reduce redundancy.

Given a query scan, NeuroCBIR retrieved the top-$k$ most similar scans from the reference database. Let $d_i \in \{\text{CN}, \text{MCI}, \text{AD}\}$ denote the diagnosis of the $i$-th retrieved neighbor, and let $p(d)$ be the empirical class proportion in the reference dataset. To account for class imbalance, each neighbor was weighted inversely to its class proportion.

The risk score for class $c$ was computed as the weighted proportion of neighbors in that class:

\begin{equation}
    R(c) = \frac{\sum_{i=1}^{k} w_i \, \mathbf{1}_{d_i = c}}{\sum_{i=1}^{k} w_i},
\end{equation}
where $w_i = 1 / p(d_i)$ is the inverse-prior weight, and $\mathbf{1}_{d_i = c}$ is 1 if neighbor $i$ belongs to class $c$, and 0 otherwise. The predicted class is the one with the highest risk score.

We quantified classification performance using:

\begin{itemize}
    \item Balanced accuracy (BAcc): the average per-class sensitivity, accounting for class imbalance:
    \begin{equation}
        \text{BAcc} = \frac{1}{|C|} \sum_{c \in C} \frac{\text{TP}_c}{\text{TP}_c + \text{FN}_c}.
    \end{equation}

    \item Macro-F1 score (MF1): the unweighted mean of per-class F1 scores:
    \begin{equation}
        \text{MF1} = \frac{1}{|C|} \sum_{c \in C} \frac{2 \cdot \text{TP}_c}{2 \cdot \text{TP}_c + \text{FP}_c + \text{FN}_c}.
    \end{equation}
\end{itemize}

\section{Results}
\label{sec:results}

\subsection{Hyperparameter optimization}
\label{subsec:results_hyp_op}
Table~\ref{tab:hyper_opt} summarizes the hyperparameter optimization results for the whole-brain retrieval model. Performance was primarily assessed using mAP@5, reflecting short-range retrieval accuracy, while embedding-level perceptual alignment was evaluated via Spearman’s rank correlation with MS-SSIM ($\rho_{\mathrm{MS\text{-}SSIM}}$). Across configurations, higher retrieval performance generally coincided with stronger (more negative) correlations, indicating improved consistency between embedding similarity and perceptual image similarity. The final configuration was selected based on the highest mAP@5 score, corresponding to $N_f = 32$, $\alpha_{\mathrm{hn}} = 0.2$, $\alpha_{\mathrm{sn}} = 0.2$, $\lambda_{\mathrm{sn}} = 0.2$, and $\pi_{\mathrm{hn}} = 5$. This setting was used for all subsequent whole-brain experiments. To ensure a fair and consistent comparison, the same hyperparameter configuration was also adopted for the region-based retrieval approach. 

\reviadd{Moreover, we performed three ablation experiments by setting $\lambda_{\mathrm{sn}}=0$, $\lambda_{\mathrm{hn}}=0$, or both (last three rows of Table~\ref{tab:hyper_opt}). The results indicate distinct roles for the two loss terms. Removing the soft-negative loss markedly reduces retrieval performance (mAP@5: 98.6\% to 79.0\%), whereas removing the hard-negative loss preserves high retrieval accuracy but substantially weakens the agreement with perceptual similarity ($\rho_{\mathrm{MS-SSIM}}$: $-0.71$ to $-0.01$). Although removing both loss terms yields the highest mAP@5, the resulting embedding is considerably less consistent with MS-SSIM, supporting the inclusion of both losses in the final configuration.}

\begin{table}[!ht]
\begin{threeparttable}
\caption{Hyperparameter optimization results for the whole-brain approach. \reviadd{For the validation partition:} retrieval quality is assessed using mAP@5, while embedding quality is evaluated via Spearman’s rank correlation between embedding similarities and MS-SSIM ($\rho_{\mathrm{MS-SSIM}}$). \reviadd{The last three rows correspond to ablations of the selected model ($\star$).}}
\label{tab:hyper_opt}
\centering
\small
\begin{tabular*}{\columnwidth}{@{\extracolsep{\fill}}S[table-format=3,round-mode=places,round-precision=0]S[table-format=1.1]S[table-format=1.2]S[table-format=1.2]S[table-format=2,round-mode=places,round-precision=0]S[table-format=1.2]S[table-format=2.1]l}
\toprule
\text{$\mathbf{N_{f}}$} & 
\text{$\boldsymbol{\alpha}_{\mathbf{hn}}$} & 
\text{$\boldsymbol{\lambda}_{\mathbf{sn}}$} &
\text{$\boldsymbol{\alpha}_{\mathbf{sn}}$} & 
\text{$\boldsymbol{\pi}_{\mathbf{hn}}$} &
\text{$\boldsymbol{\lambda_{\mathrm{hn}}}$} &
\text{\textbf{\reviadd{mAP@5\textsuperscript{$\uparrow$}}}} & 
\text{\reviadd{$\boldsymbol{\rho}_{\mathbf{MS-SSIM}}$\textsuperscript{$\downarrow$}}} \\
\midrule

16  & 0.5   & 1.0    & 0.05  & 30  & 0.10    & 94.7  & $-0.72$    \\
16  & 0.2   & 1.0    & 0.05  & 30  & 0.10    & 95.3  & $-0.71$     \\
32  & 0.5   & 1.0    & 0.05  & 30  & 0.10    & 95.2  & $-0.70$     \\
32  & 0.2   & 1.0    & 0.05  & 30  & 0.10    & 96.9  & $-0.74$     \\
32  & 0.2   & 0.2    & 0.05  & 30  & 0.10    & 97.0  & $\mathbf{-0.74}$     \\
32  & 0.2   & 0.2    & 0.10  & 30  & 0.10    & 97.6  & $-0.73$     \\
32  & 0.2   & 0.2    & 0.10  & 5   & 0.10    & 98.3  & $-0.72$      \\
32  & 0.2   & 0.2    & 0.20  & 5   & 0.10    & 98.6  & $-0.71$\hspace{3mm}$\star$ \\
64  & 0.5   & 1.0    & 0.05  & 30  & 0.10    & 96.4  & $-0.71$     \\
64  & 0.2   & 0.2    & 0.10  & 5   & 0.10    & 98.0  & $-0.71$      \\
128 & 0.5   & 1.0    & 0.05  & 30  & 0.10    & 96.2  & $-0.70$    \\
256 & 0.5   & 1.0    & 0.05  & 30  & 0.10    & 95.6  & $-0.70$     \\
\midrule
32  & 0.2   & 0.0    & 0.20  & 5   & 0.10    & 79.0  & $-0.70$      \\
32  & 0.2   & 0.2    & 0.20  & 5   & 0.00    & 93.3  & $-0.01$      \\
32  & 0.2   & 0.0    & 0.20  & 5   & 0.00    & \textbf{99.6}  & $-0.34$      \\

\bottomrule
\end{tabular*}
    \begin{tablenotes}
        {\footnotesize
        \setlength{\leftmargin}{0pt}
        \setlength{\labelsep}{0pt}
        \item[] Note: The abbreviations used in the table are as follows:
        $\mathrm{N_f}$ is the embedding dimensionality; 
        $\alpha_{\mathrm{hn}}$ is the scaling factor of the hard-negative margins; 
        $\alpha_{\mathrm{sn}}$ is the scaling factor of the soft-negative margins; 
        $\lambda_{\mathrm{hn}}$ controls the contribution of the hard-negative loss term; 
        $\pi_{\mathrm{hn}}$ is the lower percentile threshold;
        and $\pi\mathrm{_{sn}}=0.95$ is the upper percentile threshold.
        \reviadd{\textsuperscript{$\downarrow$}~indicates lower values are better.
        \textsuperscript{$\uparrow$}~indicates higher values are better.
        $\star$ denotes the configuration selected for all subsequent experiments and the ablation study.}

        }
        
    \end{tablenotes}
\end{threeparttable}
\end{table}

\subsection{Re-identification and embedding-space analysis}

\begin{figure*}[!ht]
  \centering
  \includegraphics[width=\textwidth]{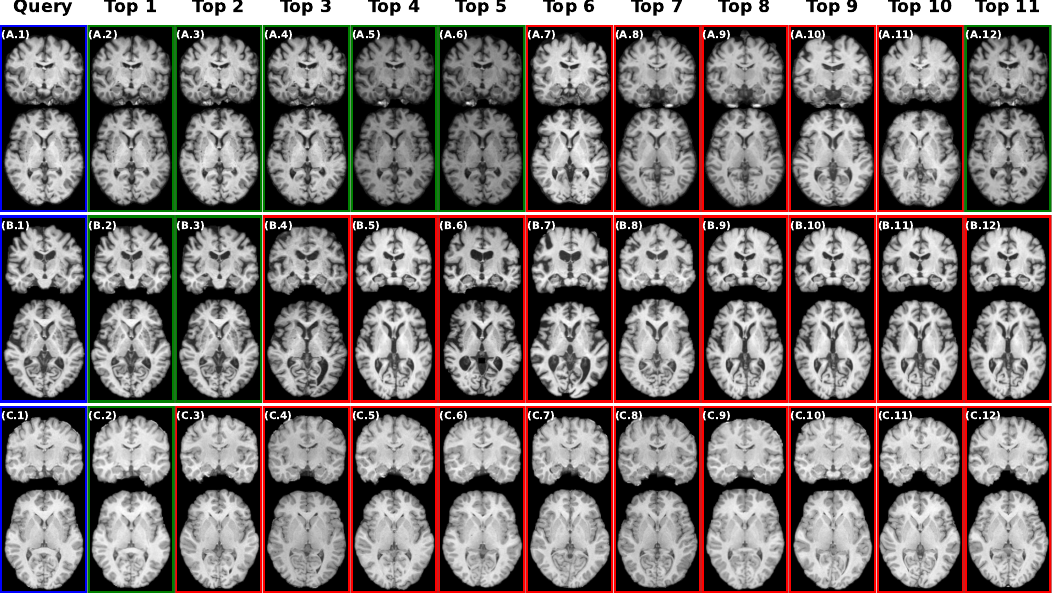}
  \caption{Retrieval examples of T1w brain MRI scans. A.1 belongs to a subject with 7 MRI scans. B.1 are from a subject with only 3 scans. C.1 belongs to a patient with only two scans. The blue frame denotes the queries, the green frames are for the "hits" (same subject scans), and the red frames are for the different subject scans that are deemed to be similar to the query by NeuroCBIR. Except for A.12, images from the same subject are ranked higher by NeuroCBIR.}
  \label{fig:example_brain_retrieval}
\end{figure*}

\begin{table*}
\begin{threeparttable}
    \caption{Retrieval re-identification performance (mAP@5, mS@1, mS@5) and Spearman’s rank correlation between embedding similarities and MS-SSIM ($\rho_{\mathrm{MS-SSIM}}$) for the whole-brain NeuroCBIR framework across multiple datasets.}
    \label{tab:wb_cbir_metrics}
    \centering
    \begin{tabular*}{\textwidth}{>{\centering\arraybackslash}p{2.5cm}@{\extracolsep{\fill}}S[table-format=3.2]S[table-format=3.2]S[table-format=3.2]S[table-format=3.2]S[table-format=3.2]S[table-format=3.2]S[table-format=3.2]S[table-format=3.2]S[table-format=3.2]}
        \toprule
        \multicolumn{1}{c}{\textbf{Metric}} &
        \multicolumn{1}{c}{\textbf{All}} &
        \multicolumn{1}{c}{\textbf{Train}} &
        \multicolumn{1}{c}{\textbf{Val}} &
        \multicolumn{1}{c}{\textbf{Test}} &
        \multicolumn{1}{c}{\textbf{ADNI}} &
        \multicolumn{1}{c}{\textbf{OASIS3}} &
        \multicolumn{1}{c}{\textbf{AIBL}} &
        \multicolumn{1}{c}{\textbf{MIRIAD}} &
        \multicolumn{1}{c}{\textbf{SLIM}\tnote{1}} \\
        \midrule
        \reviadd{mAP@5} ($\uparrow$)                           & 99.2      & 99.6      & 98.7     & 98.4     & 99.3      & 92.5     & 100.0   & 99.9  & \text{-} \\
        \reviadd{mS@1} ($\uparrow$)                            & 98.8      & 99.4      & 99.0     & 97.2     & 99.9      & 93.6     & 94.9    & 100.0   & 89.1     \\
        \reviadd{mS@5} ($\uparrow$)                            & 99.9      & 100.0     & 99.8     & 99.8     & 99.9      & 99.4     & 100.0   & 100.0  & \text{-} \\
        \reviadd{$\rho_{\mathrm{MS-SSIM}}$} ($\downarrow$)       & -0.71 & -0.64 & -0.71 & -0.74 & -0.71 & -0.65 & -0.75 & -0.77  & -0.79 \\
        \bottomrule

    \end{tabular*}
    \begin{tablenotes}
        {\footnotesize
        \item [1] The SLIM dataset contains only 3 scans per subject, so mAP@5 and mS@5 are not available.}
     \end{tablenotes}
\end{threeparttable}
\end{table*}
Table~\ref{tab:wb_cbir_metrics} shows that the whole-brain NeuroCBIR framework achieves consistently high retrieval performance across testing query datasets, with mAP and mS exceeding 95\% for most cases. Performance is near-perfect on the training set and remains robust on both the test subset and three external datasets, indicating strong generalization. The smaller datasets, such as AIBL and SLIM, show slightly lower mS@1 due to limited longitudinal sample data per subject and the lack of repeated scans per subject, which makes retrieval more challenging. By contrast, MIRIAD includes multiple scans per subject at similar time points, facilitating retrieval and explaining its higher scores despite a similar sample size. In contrast, mS@5 consistently approaches 100\% across datasets, confirming the reliability of the system for retrieving highly relevant matches. In Figure~\ref{fig:example_brain_retrieval}, each row shows a queried whole-brain MRI and its top-5 most similar retrieved scans.

\begin{figure}[!tb]
  \centering
  \small
  \includegraphics{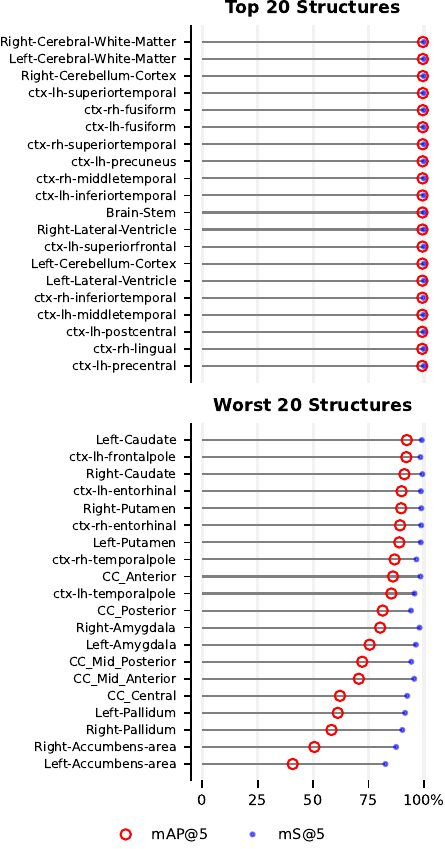}
  \caption{mAP@5 and mS@5 for the testing partition of the top 20 best / worst brain structures sorted by mAP@5. Abbreviations: CC = Corpus Callosum; Mid = Middle; lh/rh = left/right hemisphere; ctx = cortex.}
  \label{fig:top_20_structures}
\end{figure}

Figure~\ref{fig:top_20_structures} and Table~\ref{tab:br_cbir_metrics} show the retrieval performance of the brain-region NeuroCBIR framework across multiple datasets and anatomical structures. For clarity, Table~\ref{tab:br_cbir_metrics} reports results for only four representative structures (Left-Hippocampus, Left-Thalamus, Left-Amygdala, and Left-Lateral-Ventricle) which span the range of performance observed in Figure~\ref{fig:top_20_structures}. The full detailed results are provided in~\ref{app:detailed_region_metrics}. Overall, the system exhibits high precision and success, particularly for large and well-defined regions such as the hippocampus and lateral ventricles, where mS@1 precision exceeds 98\% in most datasets. 

\begin{table*}[!ht]
\begin{threeparttable}
    \caption{
    Retrieval re-identification performance (mAP@5, mS@1, mS@5) and Spearman’s rank correlation between embedding similarities and MS-SSIM ($\rho_{\mathrm{MS-SSIM}}$) for different brain-regions of the NeuroCBIR framework across multiple datasets. The full detailed results are provided in~\ref{app:detailed_region_metrics}.}
    \label{tab:br_cbir_metrics}
    \centering
    \small
    \begin{tabular*}{\textwidth}{c@{\extracolsep{\fill}}lS[table-format=3.2]S[table-format=3.2]S[table-format=3.2]S[table-format=3.2]S[table-format=3.2]S[table-format=3.2]S[table-format=3.2]S[table-format=3.2]S[table-format=3.2]}

    \toprule
    \multicolumn{1}{c}{\textbf{Metric}} & 
    \multicolumn{1}{c}{\textbf{Structure}} & 
    \multicolumn{1}{c}{\textbf{\reviadd{All}}} & 
    \multicolumn{1}{c}{\textbf{\reviadd{Train}}} & 
    \multicolumn{1}{c}{\textbf{\reviadd{Val}}} & 
    \multicolumn{1}{c}{\textbf{\reviadd{Test}}} & 
    \multicolumn{1}{c}{\textbf{\reviadd{ADNI}}} & 
    \multicolumn{1}{c}{\textbf{\reviadd{OASIS3}}} & 
    \multicolumn{1}{c}{\textbf{\reviadd{AIBL}}} & 
    \multicolumn{1}{c}{\textbf{\reviadd{MIRIAD}}} & 
    \multicolumn{1}{c}{\textbf{\reviadd{SLIM}}\tnote{1}} \\
    \midrule
    \multirow{4}{*}{mAP@5 ($\uparrow$)} 
                        & Left-Hippocampus        & 98.9      & 99.3      & 98.1     & 98.3     & 99.0      & 92.7     & 100.0   & 100.0    & \text{-}     \\
                        & Left-Thalamus           & 96.8      & 98.5      & 93.2     & 93.6     & 97.0      & 81.5     & 100.0   & 99.5    & \text{-}     \\
                        & Left-Amygdala           & 86.0      & 91.4      & 73.8     & 75.5     & 85.8      & 72.2     & 96.7    & 96.3    & \text{-}     \\
                        & Left-Lateral-Ventricle  & 99.6      & 99.7      & 99.4     & 99.3     & 99.6      & 96.0     & 100.0   & 100.0    & \text{-}     \\
                        \midrule
    \multirow{4}{*}{mS@1 ($\uparrow$)} 
                        & Left-Hippocampus        & 98.8      & 99.1      & 99.1     & 97.7     & 99.7      & 92.5     & 97.5    & 100.0   & 91.6 \\
                        & Left-Thalamus           & 95.9      & 97.5      & 96.7     & 91.5     & 98.7      & 80.6     & 88.0    & 100.0   & 69.3 \\
                        & Left-Amygdala           & 88.0      & 91.4      & 85.5     & 80.5     & 91.7      & 65.0     & 75.3    & 99.3    & 57.5 \\
                        & Left-Lateral-Ventricle  & 99.5      & 99.8      & 99.7     & 98.8     & 99.9      & 98.4     & 97.7    & 100.0   & 94.3 \\
                        \midrule
    \multirow{4}{*}{mS@5 ($\uparrow$)} 
                        & Left-Hippocampus        & 99.8      & 99.9      & 99.8     & 99.8     & 99.9      & 99.1     & 100.0   & 100.0    & \text{-}     \\
                        & Left-Thalamus           & 99.7      & 99.9      & 99.6     & 99.4     & 99.8      & 96.7     & 100.0   & 100.0    & \text{-}     \\
                        & Left-Amygdala           & 97.8      & 98.4      & 96.8     & 96.3     & 97.9      & 90.2     & 100.0   & 100.0    & \text{-}     \\
                        & Left-Lateral-Ventricle  & 100.0     & 100.0     & 99.9     & 99.9     & 100.0     & 99.7     & 100.0   & 100.0   & \text{-}     \\
                        \midrule
    \multirow{4}{*}{$\rho_{\mathrm{MS-SSIM}}$ ($\downarrow$)} 
                        & Left-Hippocampus        & -0.76 & -0.75 & -0.75 & -0.77 & -0.75 & -0.75 & -0.76 & -0.8   & -0.8  \\
                        & Left-Thalamus           & -0.75 & -0.7  & -0.74 & -0.78 & -0.73 & -0.72 & -0.78 & -0.78  & -0.82 \\
                        & Left-Amygdala           & -0.74 & -0.73 & -0.75 & -0.75 & -0.75 & -0.72 & -0.77 & -0.77  & -0.75 \\
                        & Left-Lateral-Ventricle  & -0.71 & -0.69 & -0.67 & -0.74 & -0.63 & -0.75 & -0.7  & -0.72  & -0.85 \\
    \bottomrule
    \end{tabular*}%
    \begin{tablenotes}
        {\footnotesize
        \item [1] The SLIM dataset contains only 3 scans per subject, so mAP@5 and mS@5 are not available.}
     \end{tablenotes}
\end{threeparttable}
\end{table*}

\begin{figure}[!ht]
  \centering
  \includegraphics[width=\columnwidth]{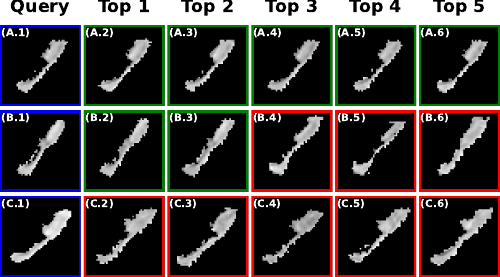}
  \caption{Retrieval of similar left hippocampus segmentations. A.1 belongs to a subject with at least 6 MRI scans. B.1 are from a subject with only 5 scans (B1.1--B1.5). C.1 belongs to a patient with no additional MRI scans. The blue frame denotes the queries, the green frames are for the "hits", and the red frames are for the "fails".}
  \label{fig:example_struct_retrieval}
\end{figure}

The drop in performance for certain brain regions in Table~\ref{tab:br_cbir_metrics} could be associated with multiple reasons: (i) the extra postprocessing step required for segmenting these regions. (ii) Because of the limited longitudinal samples per subject (e.g., AIBL and SLIM), it highlights the influence of dataset size and variability on retrieval performance. (iii) ADNI and MIRIAD datasets contain repeated measures of the same scan for the same session, which is reflected in the higher performance with respect to OASIS3, AIBL, and SLIM, which only contain an MRI scan per session. (iv) Additionally, small or homogeneous structures such as the amygdala and thalamus may not exhibit overtly distinct individual patterns, resulting in lower retrieval performance. Notably, increasing the retrieval rank from mS@1 to mS@5 consistently improves, approaching 100\% for most structures, which demonstrates the robustness of the embeddings in reliably retrieving relevant matches. Figure~\ref{fig:example_struct_retrieval} demonstrates retrieval at the level of individual brain structures, using left hippocampus segmentations as examples. 

\subsection{Interpretability evaluation}
\begin{figure}[!ht]
  \centering
  \includegraphics{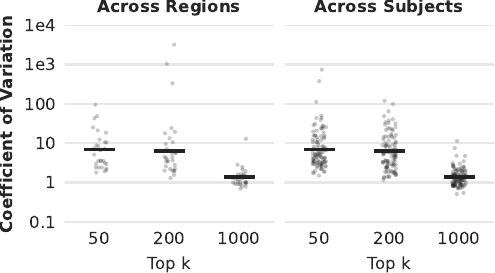}
  \caption{Coefficient of Variation (CV) of Spearman correlations between whole-brain similarity and region-based similarity across top-$k$ retrieved cases, excluding scans from the query subject. Left panel shows variability across regions for each query, and right panel shows variability across queries for each region. Top-$k$ groups of 50, 200, and 1000 are shown. The CV is plotted on a logarithmic scale to highlight differences across a wide range of values. Points indicate individual observations, while horizontal markers show the mean CV for each top-$k$ group. Higher CV indicates greater heterogeneity in the contribution of regions or queries to whole-brain similarity.}
  \label{fig:cv_across_top_k}
\end{figure}

The CV of the Spearman correlations across regions and across queries highlights the relative heterogeneity of regional contributions to whole-brain similarity (Figure~\ref{fig:cv_across_top_k}). For the top-1000 retrieved neighbors across queries, several large anatomical regions exhibit low variability (CV < 1), most notably the left and right cerebral white matter and the lateral ventricles, indicating stable contributions of these regions. In contrast, higher CV values are observed in cortical regions such as the right parahippocampal, caudal anterior cingulate, entorhinal, and temporal pole cortices, reflecting greater query-dependent variability.

Across-region CV analysis further shows that, for some queries at top-1000, CV values fall below 1, suggesting relatively uniform contributions of all regions to whole-brain similarity in those cases. 

\subsection{Bias assessment}
Table~\ref{tab:comparison_results} outlines the statistical comparisons between different \reviadd{classes}, datasets, scanner field strengths, and manufacturers, including p-values and effect sizes (Cliff’s \textdelta). Most inter-dataset comparisons, such as test vs. training and ADNI vs OASIS3, showed highly significant differences (p \textless 0.001) but with negligible to small effect sizes. This finding indicates that although statistically detectable, the practical differences are limited. Comparisons involving the MIRIAD dataset generally exhibited smaller differences, reflecting its relative similarity to other datasets. \reviadd{classes}, field strengths, and manufacturer comparisons revealed minimal effects, with most Cliff’s \textdelta values close to zero, confirming that these factors had a negligible impact on distributions. Overall, the table highlights that while statistical significance is observed in many pairwise tests, the effect sizes are mostly negligible or small, suggesting robust generalization across datasets and acquisition conditions.

\begin{table}[!ht]
\begin{threeparttable}

    \caption{Statistical comparison of retrieval performance across datasets, scanner field strengths, and vendors. Reported values include \textit{p}-values (Mann–Whitney U test), significance indicators (ssd), Cliff’s $\delta$ effect sizes, and their qualitative magnitudes. Asterisks (*) denote significant differences at \textit{p}-value\textless0.05.}
    \label{tab:comparison_results}
    \centering
    \small
    \sisetup{
        round-mode=places,
        round-precision=4
    }
    \setlength{\tabcolsep}{4.3pt}
    \begin{tabular*}{\columnwidth}{@{\extracolsep{\fill}}l S[table-format=1.4] c S[table-format=-1.4] r}
        \toprule
        \textbf{Comparison} & \reviadd{\textbf{\textit{p}-value}} & \reviadd{\textbf{ssd}} & \reviadd{\textbf{Cliff’s $\boldsymbol\delta$}} & \reviadd{\textbf{Magnitude}} \\
        \midrule
        Test vs. Train       & 0.000000 & * & -0.027000 & negligible \\
        \reviadd{CN vs. MCI}       & 0.003080 & * & -0.006000 & negligible \\
        \reviadd{CN vs. AD}        & 0.685000 &   & -0.004000 & negligible \\
        \reviadd{MCI vs. AD}       & 1.000000 & * &  0.002000 & negligible \\
        ADNI vs. AIBL        & 1.000000 &  & -0.009000 & negligible \\
        ADNI vs. MIRIAD      & 0.338000 &  & -0.006000 & negligible \\
        ADNI vs. OASIS3      & 0.000000 & * & 0.152000 & small \\
        AIBL vs. MIRIAD      & 1.000000 &  & 0.003000 & negligible \\
        AIBL vs. OASIS3      & 0.863000 &  & 0.161000 & small \\
        MIRIAD vs. OASIS3    & 0.000000 & * & 0.159000 & small \\
        1.5T vs. 3T          & 0.000000 & * & 0.008000 & negligible \\
        GE vs. PH       & 0.590000 &  & 0.001000 & negligible \\
        GE vs. SH       & 0.000000 & * & 0.010000 & negligible \\
        PH vs. SH & 0.003000 & * & 0.009000 & negligible \\
        \bottomrule
    \end{tabular*}
    \begin{tablenotes}
        {\footnotesize
        \setlength{\leftmargin}{0pt}
        \setlength{\labelsep}{0pt}
        \item[] Note: The abbreviations used in the table are as follows:
        Siemens Healthineers (SH), General Electric (GE) and Philips Healthcare (PH).
        }
     \end{tablenotes}
\end{threeparttable}
\end{table}

\subsection{Comparison with pretrained models}
Table~\ref{tab:comparison_models} compares retrieval performance across different feature extractors and embedding strategies. While standard CNN backbones and pretrained vision models achieve moderate retrieval accuracy, NeuroCBIR markedly outperforms all baselines, reaching a mAP@5 of \reviadd{98.3}\revidel{98.5}\% and a mS@5 of 99.8\% with only 32 features. Importantly, NeuroCBIR also shows a strong negative Spearman correlation with MS-SSIM ($\rho_{\mathrm{MS\text{-}SSIM}}=-0.74$), indicating close agreement between embedding-based similarity rankings and perceptual image similarity. In contrast, baseline models exhibit weaker correlations, suggesting less structured and less interpretable embedding spaces. 

\begin{table}[!ht]
\begin{threeparttable}
\caption{Comparison of different feature extractors and dimensionality reduction approaches on the retrieval task using the top-5 results from the whole-brain testing set. Reported metrics include the compression rate (Comp.), number of features ($\mathrm{N_{f}}$), mean precision, and success rate.}
\label{tab:comparison_models}
\centering
\small
\setlength{\tabcolsep}{3pt}
\begin{tabular*}{\columnwidth}{@{\extracolsep{\fill}}lrrS[table-format=2.1]S[table-format=2.1]c}
\toprule
{\textbf{Method}}     & \textbf{Comp.}     & 
$\mathbf{N_{f}}$\textsuperscript{$\downarrow$}  & 
\textbf{mAP@5}\textsuperscript{$\uparrow$} & 
\textbf{mS@5}\textsuperscript{$\uparrow$} & 
$\boldsymbol{\rho}_{\mathbf{MS-SSIM}}$\textsuperscript{$\downarrow$} \\
\midrule
ResNet-10       & \num[round-mode=places,round-precision=0]{2048}:1     & 512   & 7.2    & 32.6 & $-0.55$      \\
ResNet-50       & \num[round-mode=places,round-precision=0]{512}:1      & 2048  & 10.8   & 43.9 & $-0.70$     \\
ResNet-101      & \num[round-mode=places,round-precision=0]{512}:1      & 2048  & 5.7    & 30.0 & $-0.51$\\
CNN-AD          & \num[round-mode=places,round-precision=0]{2628}:1     & 256   & 0.1    & 0.7 & $-0.01$     \\
VoComni-L       & \num[round-mode=places,round-precision=0]{576}:1      & 1536  & 14.3   & 50.9 & $-0.34$     \\
\reviadd{BrainIAC}        & \num[round-mode=places,round-precision=0]{1152}:1     & 768  & 72.4   & 95.8 & $-0.27$     \\
\reviadd{FOMO2JOMO}       & \num[round-mode=places,round-precision=0]{1024}:1     & 256  & 80.3   & 97.2 & $-0.29$     \\
VAE+UMAP        & \num[round-mode=places,round-precision=0]{183040}:1   & \textbf{32}    & 17.4   & 57.4 & $-0.32$     \\
\reviadd{NeuroCBIR}       & \num[round-mode=places,round-precision=0]{183040}:1   & \textbf{32}    & 98.3   & 99.8  & $\mathbf{-0.74}$    \\
\reviadd{\hspace{2mm}''\hspace{2mm} ($\lambda_{\mathrm{sn}}=0$)}    & \num[round-mode=places,round-precision=0]{183040}:1   & \textbf{32}    & 79.0   & 97.2  & $-0.73$    \\
\reviadd{\hspace{2mm}''\hspace{2mm} ($\lambda_{\mathrm{hn}}=0$)}    & \num[round-mode=places,round-precision=0]{183040}:1   & \textbf{32}    & 92.8   & 99.3  & $-0.03$    \\
\reviadd{\hspace{2mm}''\hspace{2mm} ($\lambda_{\mathrm{hn}}=\lambda_{\mathrm{sn}}=0$)}    & \num[round-mode=places,round-precision=0]{183040}:1   & \textbf{32}    & \textbf{99.6}   & \textbf{100.0}  & $-0.41$    \\

\bottomrule
\end{tabular*}
\begin{tablenotes}
        {\footnotesize
        \setlength{\leftmargin}{0pt}
        \setlength{\labelsep}{0pt}
        \item[] Note: \textsuperscript{$\downarrow$} indicates lower values are better. \textsuperscript{$\uparrow$} indicates higher values are better.}
     \end{tablenotes}
\end{threeparttable}
\end{table}

\subsection{Computational complexity}
Table~\ref{tab:execution_time} summarizes the execution time and memory requirements of the NeuroCBIR system for both whole-brain and brain-region processing. As expected, compared to brain-region retrieval, whole-brain retrieval is more computationally demanding, with the Q2E forward pass requiring on average 66.4 seconds on a single core and 18.7 seconds on four cores, accompanied by a peak memory usage of 6.8~GB. In contrast, region-specific retrieval is substantially more efficient, reducing both memory consumption (peak 1.4~GB) and processing time (3.6 seconds on a single core and 1.2 seconds on four cores). In both cases, the retrieval step itself (top-5 similarity search) required negligible time (\textless0.01 seconds), confirming that the system is dominated by the embedding extraction step.

\begin{table}[!ht]
\caption{Execution time and memory usage for whole-brain and brain-region processing. Baseline refers to the RAM usage before running, peak refers to the peak of RAM measured during the execution.}
\label{tab:execution_time}
\centering
\small
\begin{tabular*}{\columnwidth}{@{\extracolsep{\fill}}l
                                 S[table-format=4,round-mode=places,round-precision=0] 
                                 S[table-format=4,round-mode=places,round-precision=0] 
                                 S[table-format=2.1,round-mode=places,round-precision=1] 
                                 S[table-format=2.1,round-mode=places,round-precision=1] 
                                 }
\toprule
\multirow{2}{*}{\textbf{Step}} & \multicolumn{2}{c}{\textbf{RAM (MB)  $\boldsymbol{(\downarrow)}$}} & \multicolumn{2}{c}{\textbf{Time (s) $\boldsymbol{(\downarrow)}$}} \\
\cmidrule{2-5}
 & {\textbf{baseline}} & {\textbf{peak}} & {\textbf{1 core}} & {\textbf{4 cores}} \\
\midrule
\textbf{Whole-brain} & & & & \\
\addlinespace[1mm]
Load Q2E        & 862  & 1087 &  8.7 &  6.5 \\
Run Q2E         & 1087 & 6804 & 66.4 & 18.7 \\
Top 5 retrieval & {-}   & {-}   &  \text{\textless0.01} & \text{\textless0.01} \\
\midrule
\textbf{Brain-regions} & & & & \\
\addlinespace[1mm]
Load Q2E        & 752  & 1000 & 0.8 & 0.8 \\
Run Q2E         & 1000 & 1394 & 3.6 & 1.2 \\
Top 5 retrieval & {-}   & {-}   & \text{\textless0.01} & \text{\textless0.01} \\
\bottomrule
\end{tabular*}
\end{table}

\subsection{Application-specific evaluation}
Next, we evaluated the performance of the embeddings in zero-shot tasks, testing whether they capture meaningful age- and disease-related patterns without supervision.

\subsubsection{Zero-shot similarity-based age inference}

Table~\ref{tab:zero_shot_age} summarizes zero-shot age estimation performance across selected brain regions using the top-$500$ retrieved cases from \reviadd{the different methods, including a random baseline with 4 random features and the ablation cases.} \revidel{NeuroCBIR}. Overall, the results indicate that similarity-based retrieval captures meaningful age-related structure in both global and regional brain representations in a fully zero-shot setting. The whole-brain approach achieves the best performance, with a MAE of 4.3 years and a Pearson correlation of $r=0.54$, outperforming all individual region-based estimates \reviadd{as well as the alternative embedding approaches evaluated in this study}.

\reviadd{The ablation results show that both loss terms contribute to the learned representations, with removing the hard-negative loss producing the largest degradation (MAE = 5.2 years, $r=0.09$), while removing the soft-negative loss resulted in a smaller decrease (MAE = 4.7 years, $r=0.49$).}

\begin{figure*}[!ht]
  \centering
  \includegraphics{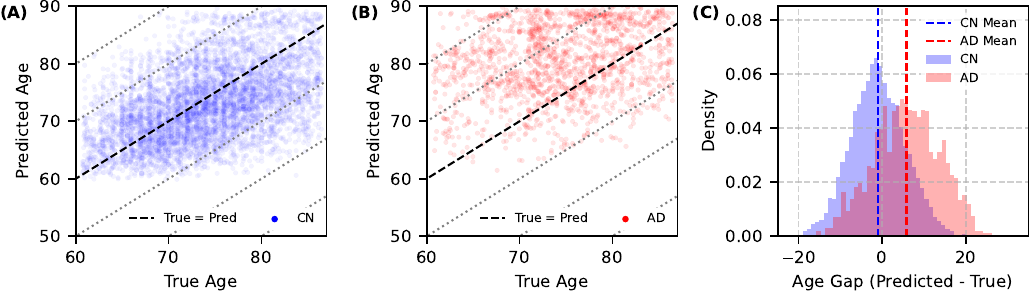}
  \caption{Zero-shot similarity-based brain age estimation using NeuroCBIR for the left hippocampus. Predicted age is obtained by aggregating the ages of the top-$500$ most similar CN subjects. Panel (A) shows predicted versus chronological age for CN queries, panel (B) for AD queries, and panel (C) compares the age-gap distributions for CN and AD.}
  \label{fig:zero_shot_age}
\end{figure*}

Figure~\ref{fig:zero_shot_age} illustrates the relationship between predicted and chronological age for the left hippocampus in CN and AD subjects. For CN queries, the age-gap distribution is centered around zero, as expected, since both the query and retrieval sets consist of CN subjects. In contrast, AD queries exhibit a positive age bias of approximately 6 years, indicating that AD scans tend to retrieve older CN subjects.

\begin{table}[!ht]
\begin{threeparttable}
\caption{Zero-shot similarity-based brain age estimation across selected brain regions. For each region, the predicted age was computed using the similarity-based weighted average of the top-$500$ retrieved embeddings. Median absolute error (MAE) in years is reported as mean ± standard deviation. Pearson correlation $r$ and regression slope indicate the correspondence between predicted and chronological age.}
\label{tab:zero_shot_age}
\centering
\small
\begin{tabular*}{\columnwidth}{@{\extracolsep{\fill}}lcccc}
\toprule
\textbf{Region} & \textbf{MAE} $\boldsymbol{(\downarrow)}$ & $\boldsymbol{r~(\uparrow)}$ & \textbf{Slope} \\
\midrule
\textbf{Whole-brain} & & & \\
\addlinespace[1mm]
\reviadd{Random baseline}         & 5.2 ± 3.5 & 0.00 & 0.00 \\
\reviadd{ResNet-10}               & 4.5 ± 3.5 & 0.46 & 0.32 \\
\reviadd{ResNet-50}               & 4.8 ± 3.6 & 0.35 & 0.16 \\
\reviadd{ResNet-101}              & 5.0 ± 3.6 & 0.27 & 0.13 \\
\reviadd{CNN-AD}                  & 5.8 ± 4.0 & 0.00 & 0.00 \\
\reviadd{VoComni-L}               & 5.0 ± 3.6 & 0.34 & 0.22 \\
\reviadd{BrainIAC}                & 5.0 ± 3.8 & 0.42 & 0.36 \\
\reviadd{FOMO2JOMO}               & 4.8 ± 3.6 & 0.48 & 0.42 \\
\reviadd{NeuroCBIR}               & \textbf{4.3 ± 3.2} & \textbf{0.54} & 0.40 \\
\reviadd{NeuroCBIR ($\lambda_{\mathrm{sn}}=0$)}               & 4.7 ± 3.4 & 0.49 & 0.39 \\
\reviadd{NeuroCBIR ($\lambda_{\mathrm{hn}}=0$)}               & 5.2 ± 3.6 & 0.09 & 0.02 \\
\reviadd{NeuroCBIR ($\lambda_{\mathrm{hn}}=\lambda_{\mathrm{sn}}=0$)}               & 4.4 ± 3.1 & 0.51 & 0.27 \\
\addlinespace[1.2mm]
\reviadd{\textit{NeuroCBIR (AIBL)}}              & \textit{4.5 ± 3.4 }& \textit{0.42} & \textit{0.25} \\
\reviadd{\textit{\citet{rajabliBrainAgePrediction2025} (AIBL)}\textsuperscript{1}}   & \textit{3.8 ± 0.4} & \textit{0.84} & {-} \\
\midrule
\textbf{\reviadd{Brain regions - NeuroCBIR}}  & & & \\
\addlinespace[1mm]
Left Hippocampus        & 4.5 ± 3.2 & 0.49 & 0.30 \\
Left Thalamus          & 4.5 ± 3.3 & 0.51 & 0.38 \\
Left Amygdala           & 4.5 ± 3.3 & 0.47 & 0.28 \\
Left Lateral Ventricle  & \textbf{4.6 ± 3.5} & \textbf{0.51} & 0.43 \\
\bottomrule
\end{tabular*}
\begin{tablenotes}
        {\footnotesize
        \setlength{\leftmargin}{0pt}
        \setlength{\labelsep}{0pt}
        \item[] \reviadd{Note: Italics denote results obtained using the AIBL query set. Bold indicates the best performance among the evaluated retrieval-based methods. \textsuperscript{1} Performance reported using a supervised brain age estimation model, included as a reference baseline.}
        }
     \end{tablenotes}
\end{threeparttable}
\end{table}

\subsubsection{Zero-shot similarity-based disease risk estimation}

Table \ref{tab:zero_shot_disease} summarizes the performance of zero-shot similarity-based disease risk estimation across brain regions using top-$500$ retrieved neighbors \reviadd{for the different tested methods, including a random baseline with 4 features and the ablation cases.} Across all regions, performance exceeds chance level (BAcc = 33\%), with BAcc ranging from 49.7\% (left thalamus) to 57.9\% (left amygdala) and MF1 scores between 45.8\% and 55.3\%. Whole-brain embeddings achieve a BAcc of 56.3\% and a MF1 of 53.1\%, \reviadd{outperforming the alternative embedding approaches evaluated in this study}.

\reviadd{The ablation results indicate that both the soft-negative and hard-negative losses contribute to the quality of the learned embeddings. Removing the soft-negative loss reduced the three-class balanced accuracy from 56.3\% to 54.1\%, whereas removing the hard-negative loss caused a much larger drop to 38.4\% (binary: 76.7\% to 55.4\%), demonstrating the importance of both objectives, particularly the hard-negative term.}

\begin{figure*}[!ht]
  \centering
  \includegraphics{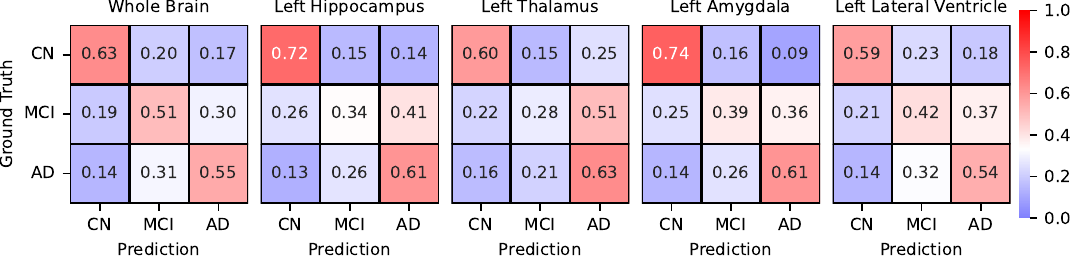}
  \caption{Row-normalized confusion matrices for zero-shot similarity-based disease risk estimation across brain regions (Top-k = 300). Rows are normalized by true class, showing per-class recall (chance-level performance corresponds to 0.33.}
  \label{fig:confusion_matrices_row_top_300}
\end{figure*}

Figure \ref{fig:confusion_matrices_row_top_300} presents the row-normalized confusion matrices, providing insight into class-specific behavior. Whole-brain representations show comparatively improved MCI recognition relative to individual subcortical regions, with the left hippocampus and amygdala showing strong recall for CN and AD, but increased confusion for MCI cases.

\begin{table}[!ht]
\begin{threeparttable}
\caption{Zero-shot similarity-based disease risk estimation across selected brain regions. For each query, the predicted class was obtained from the similarity-weighted labels of the top-$500$ retrieved embeddings. Performance is reported for both the three-class (CN/MCI/AD) and binary (CN/AD) classification tasks using balanced accuracy (BAcc) and macro F1-score (MF1).}
\label{tab:zero_shot_disease}
\centering
\small
\begin{tabular*}{\columnwidth}{@{\extracolsep{\fill}}lcccc}
\toprule
\multirow{2}{*}{\textbf{Method}} &
\multicolumn{2}{c}{\textbf{CN/MCI/AD}} & 
\multicolumn{2}{c}{\textbf{CN/AD}} \\
\cmidrule{2-5}
 & 
\textbf{BAcc\textsuperscript{$\uparrow$}} & \textbf{MF1\textsuperscript{$\uparrow$}} &
\textbf{BAcc\textsuperscript{$\uparrow$}} & \textbf{MF1\textsuperscript{$\uparrow$}} \\
\midrule
\textbf{Whole-brain} &  & \\
\addlinespace[1mm]
\reviadd{Random baseline}         & 33.3 & 33.3 & 50.0 & 50.0 \\
\reviadd{ResNet-10}               & 49.8 & 48.1 & 69.5 & 59.9 \\
\reviadd{ResNet-50}               & 50.2 & 48.0 & 67.8 & 60.2 \\
\reviadd{ResNet-101}              & 47.4 & 44.7 & 65.0 & 56.4 \\
\reviadd{CNN-AD}                  & 34.9 & 32.8 & 49.2 & 47.4 \ \\
\reviadd{VoComni-L}               & 48.6 & 46.1 & 66.6 & 58.0  \\
\reviadd{BrainIAC}                & 51.2 & 48.7 & 71.0 & 62.3 \\
\reviadd{FOMO2JOMO}               & 52.0 & 48.4 & 73.9 & 66.7 \\
\reviadd{NeuroCBIR}               & \textbf{56.3} & \textbf{53.2} & \textbf{76.7} & \textbf{69.1}\\
\reviadd{NeuroCBIR ($\lambda_{\mathrm{sn}}=0$)}  & 54.1 & 52.3 & 73.6 & 67.5 \\
\reviadd{NeuroCBIR ($\lambda_{\mathrm{hn}}=0$)}  & 38.4 & 37.1 & 55.4 & 49.5 \\
\reviadd{NeuroCBIR ($\lambda_{\mathrm{hn}}=\lambda_{\mathrm{sn}}=0$)} & 52.2 & 47.2 & 75.1 & 64.0 \\
\addlinespace[1.2mm]
\reviadd{\textit{NeuroCBIR (AIBL)}}              & \textit{48.6} & \textit{42.9} & \textit{75.2} & \textit{65.0} \\
\reviadd{\textit{\citet{wen_convolutional_2020} (AIBL)}\textsuperscript{1}}          &  {-}  & {-} & \textit{84.0} & {-} \\
\midrule
\textbf{\reviadd{Brain regions - NeuroCBIR}} &  & \\
\addlinespace[1mm]
Left Hippocampus        & 55.6 & 52.2 & 79.5 & 73.1 \\
Left Thalamus           & 50.2 & 45.5 & 72.6 & 63.1 \\
Left Amygdala           & \textbf{57.9} & \textbf{55.2} & \textbf{81.9} & \textbf{76.3} \\
Left Lateral Ventricle  & 51.4 & 48.2 & 72.3 & 64.3\\
\bottomrule
\end{tabular*}
\begin{tablenotes}
        {\footnotesize
        \setlength{\leftmargin}{0pt}
        \setlength{\labelsep}{0pt}
        \item[] \reviadd{Note: Italics denote results obtained using the AIBL query set. Bold indicates the best performance among the evaluated retrieval-based methods. \textsuperscript{$\uparrow$} indicates that higher values are better. \textsuperscript{1} Performance reported using a supervised classifier, included as a reference baseline.}
        }
     \end{tablenotes}
\end{threeparttable}
\end{table}

\section{Discussion}

In this study, we demonstrated that NeuroCBIR achieves high retrieval (re-identification) performance at both the whole-brain (mAP@5~$\sim$98.4) and brain-region levels, with most anatomical structures showing consistently strong results. The Spearman’s correlation between the embedding-space and MS-SSIM ($\rho_{\mathrm{MS-SSIM}}$) further confirms that the learned embeddings preserve meaningful perceptual and semantic similarities between inter-subject scans. These results indicate that the embeddings capture anatomically meaningful patterns that are consistent not only within subjects but also across different individuals. Evaluation across multiple large-scale datasets, varying field strengths, and scanner manufacturers further demonstrates that the framework generalizes robustly under heterogeneous acquisition conditions.

Traditional CBIR systems, particularly for brain MRI, have been predominantly evaluated based on their ability to retrieve images sharing the same class labels, such as tumor types or diagnostic categories like CN vs. AD  \citep{huang_retrieval_2012, Onga2019, nishimaki_loc-vae_2022, muraki_isometric_2024, tobari_domain-invariant_2025}. However, class labels provide only coarse population-level grouping and do not reflect fine-grained inter-individual neuroanatomical variability. Similarly, metrics like SSIM or other pixel-level similarity measures are often avoided as ground truth because they are highly sensitive to registration errors, intensity variations, and scanner-specific artifacts, which can confuse retrieval in large multi-cohort datasets \citep{dohmen_similarity_2025, renieblas_structural_2017}. While effective for image registration or quality assessment, such metrics emphasize local intensity correspondence rather than higher-level anatomical structure. Consequently, both class-based and pixel-level similarity metrics fail to capture subtle, subject-specific anatomical patterns that are crucial for meaningful brain MRI retrieval. 

We address these limitations by using same-subject images as ground truth for similarity, guiding the model to focus on subject-specific global/regional brain structure and pulling corresponding scans closer in the embedding space. At the same time, we leverage a VAE-trained latent space that captures fine-grained anatomical patterns across subjects, enabling anatomically similar scans to cluster while dissimilar ones are pushed apart. Learning compact latent representations with autoencoders, including variational autoencoders, has been widely adopted in CBIR systems to support similarity-based retrieval beyond hand-crafted features \citep{rafiei_class-specific_2023, wang_image_2004}.
\reviadd{The ablation analysis provides further insight into the role of the contrastive objectives. In agreement with previous contrastive learning studies showing that negative sample selection strongly influences representation quality \citep{Chen2020b, khoslaSupervisedContrastiveLearning2020}, our results indicate that the hard-negative and soft-negative terms contribute differently. The hard-negative term primarily improves the alignment between embedding similarity and anatomical similarity, increasing the correlation with MS-SSIM ($\rho_{\mathrm{MS\text{-}SSIM}}$: $-0.01$ to $-0.71$). The soft-negative term improves retrieval robustness by refining similarity relationships, encouraging anatomically similar brains to lie closer to the query's positive samples while maintaining separation from hard negatives (mAP@5: $79.0\%$ to $97.0\%$).}

The publication most closely related to our work is that of \citet{puglisi_deepbrainprint_2024}, which addresses the task of locating previous scans of the same patient, commonly referred to as subject re-identification. They proposed DeepBrainPrint, a framework that generates subject-specific fingerprints from 2D T1-weighted MR images. Their evaluation, conducted on ADNI and a synthetic benchmark comprising 142 images, defined retrieval success as returning scans from the same subject among the top-ranked results. Reported performance was high, with mAP@3 values of 95.5\% on ADNI and 91\% on the synthetic dataset \citep{puglisi_deepbrainprint_2024}. While this work demonstrates the feasibility of subject-level MRI retrieval, our approach differs in both scope and methodology. First, we extend retrieval to large-scale datasets comprising thousands of 3D T1-weighted MRIs, explicitly evaluating generalization across scanners and acquisition protocols. Second, beyond enforcing subject-level similarity, our latent space is jointly structured to preserve inter-subject anatomical relationships, enabling the learned features to support downstream tasks such as zero-shot age regression and pathology classification. Finally, our region-based formulation allows retrieval to focus on anatomically localized patterns when regional effects dominate over global structure, facilitating task-specific stratification in subsequent analyses.

Through the evaluation of NeuroCBIR on unsupervised tasks, including zero-shot age prediction and pathology stratification, we aim to demonstrate that the learned embedding features encode both age-related and disease-related anatomical patterns. To the best of our knowledge, there are no prior studies that perform brain age estimation purely through unsupervised, zero-shot, retrieval-based embeddings without supervised age labels during training.  Our method achieves an MAE of  $\sim$ $4.3$ years and a Pearson correlation of $r \sim$ 0.5 across the evaluated brain regions. While our method is not tailored for age prediction, it achieved competitive results compared to supervised brain age regression models reported in the literature. For example, in \citep{Dartora_deep_2024}, we achieved MAEs between 2.67 and 3 years with a regression model optimized for brain age. \reviadd{Similarly, on the external AIBL cohort, NeuroCBIR achieved an MAE of $4.5$ years, compared with $3.8$ years reported by the supervised brain age estimation model of \citet{rajabliBrainAgePrediction2025}.} 
Brain age estimation methods use standard interpretability methods, such as gradCAM \citep{Dartora_deep_2024, rajabliBrainAgePrediction2025}.
Instead of competing with brain age regression methods, NeuroCBIR can be used to add interpretability. Indeed, since our framework estimates age directly from the top-$k$ most similar brains in the database, it is possible to perform fine-grained analysis by inspecting differences in age between the query and retrieved scans and possibly detecting outliers. 

Additionally, consistent with observations in supervised brain age models \citep{soumya_kumari_review_2024,Dartora_deep_2024}, we also report an AD-related age gap when predictions are based on CN reference datasets (see Figure~\ref{fig:zero_shot_age}). This shift reflects disease-related morphological patterns that resemble accelerated normative aging, despite the absence of explicit disease supervision during training. 

Regarding zero-shot pathology prediction, whole-brain embeddings achieve a balanced accuracy of 56.3\% and a MF1 score of 53.1\% across CN, MCI, and AD. 
Most misclassifications between CN and AD occur toward the MCI class, reflecting its intermediate clinical status and supporting the biological plausibility of the learned embedding space. 
\revidel{In comparison, supervised deep learning models trained on labeled MRI scans generally achieve higher performance, with AD vs CN balanced accuracies > 70\% using CNN-based architectures}
\reviadd{On the external AIBL cohort, NeuroCBIR achieved balanced accuracies of $75.2\%$ using whole-brain embeddings and $81.9\%$ using the left amygdala for the CN/AD classification task, without any retraining or domain adaptation. These results approach the $84.0\%$ reported by the supervised classifier of
\citet{wen_convolutional_2020}, which is slightly better than their results on ADNI data \citep{Persson2026}.} 
Prior unsupervised approaches have also explored AD detection via clustering of MRI features \citep{inguanzo_mri_2023} or anomaly detection with autoencoder-based representations \citep{kim_anomaly_2026}, showing that disease-related patterns can be captured without labeled training data. Importantly, our framework demonstrates zero-shot multi-class classification across CN, MCI, and AD using fully unsupervised embeddings, showing that meaningful disease-related structure can be captured without any training with task-specific labels.

Researchers from the Division of Clinical Geriatrics at Karolinska Institutet created TheHiveDB \citep{muehlboeck_thehivedb_2014}, a system that allows users to retrieve images using specific search criteria (e.g., diagnosis, age, sex, biomarkers, memory tests, scanner, etc.). \reviadd{Notice that we decided not to consider any additional variable in the training of NeuroCBIR beyond image similarity to ease its integration into systems such as The HiveDB.} NeuroCBIR can be integrated into TheHiveDB or other search systems to add an extra dimension of image similarity, which could be used to answer different research questions \reviadd{by filtering the results from NeuroCBIR with additional criteria}. 
Thus, with NeuroCBIR, users could evaluate images alongside associated clinical records. 
Moreover, new scans can be added directly to the database without retraining the model, enabling the system to continuously improve coverage and utility as additional data becomes available. \reviadd{Regarding pure re-identification, NeuroCBIR can be used for detecting images from the same subject that, due to human errors, were assigned to different subject IDs, something that is not uncommon in ADNI, for example.} The relatively good performance of NeuroCBIR on zero-shot tasks of brain age and classification points to a correlation between image similarity, age, and condition, which is interesting to further study in the future.

A central motivation for adopting latent representations from VAEs rather than raw 3D brain MRIs lies in both computational efficiency and semantic compactness. Direct processing of volumetric scans is prohibitively memory-intensive: for example, training the VAE with a batch size of just one whole-brain MRI required $\sim$ 63 GB of GPU memory in our experiments. Once trained, the encoder alone requires  $\sim$ 20 GB to process a single image (batch size = 1). By precomputing the latent embeddings $\boldsymbol{\mu}$, the input size is reduced from [160, 176, 208] voxels to [8, 20, 22, 26], representing a drastic compression of the 3D volume. This reduction makes it possible to perform contrastive training with significantly larger batch sizes that would otherwise be infeasible. Such efficiency is crucial, since contrastive learning methods rely on large and diverse batches to achieve robust representations \citep{cheng_breaking_nodate}.

Despite these contributions, our study has several limitations. First, we restricted our analysis to T1w MRI. Thus, retraining would be needed for other sequences. 
Second, the approach relies on preprocessing steps, particularly skull-stripping, that are essential to prevent the embeddings from capturing face-related rather than brain-related patterns. These steps add additional time and complexity to a clinical workflow. New fast and accurate deep learning-based methods (e.g., \citet{hoopes_synthstrip_2022}) could reduce that issue. 
Moreover, although the system was evaluated on data preprocessed with pipelines slightly different from the training data, it remains unclear how robust the model is to substantially different preprocessing strategies. 

Future work should address these limitations by extending the framework to other clinically relevant brain MRI sequences, such as FLAIR scans, streamlining preprocessing, and exploring methods to improve the transparency and explainability of latent representations. Additionally, the inclusion of additional cohorts could expand the dataset of precomputed features and enable a more comprehensive use for both clinical and research applications.  \citet{wachinger_brainprint_2015} used cortical surfaces for subject re-identification using FreeSurfer. At a local scale, the shape of the hippocampus can also be potentially used for re-identification \citep{dekraker_automated_2023}. Thus, an interesting avenue to complement NeuroCBIR is to extract features from cortical surfaces. For usability, this approach will require fast and accurate methods for surface mesh extraction. However, despite promising results in the last few years (e.g., \citet{hao_surfnet_2025,BONGRATZ2024103093}), this remains a difficult problem. 

This study primarily focuses on the development and technical validation of the NeuroCBIR framework. While the method was evaluated for scan re-identification and explored in zero-shot clinical scenarios, a more comprehensive investigation of these and additional clinical applications is warranted. \reviadd{Indeed, the relatively good performance of NeuroCBIR on zero-shot tasks of brain age and classification suggests a correlation among image similarity, age, and condition, which is worth further study.} Future work should assess its potential for finer-grained tasks, such as the identification and stratification of dementia subtypes \citep{inguanzo_mri_2023}. In addition, NeuroCBIR could serve as a perceptual and reasoning component within future AI systems: its embeddings and similarity-based retrieval may inform clinical or research decision-making, such as cohort selection or patient risk prioritization, providing a foundation for more agentic AI architectures with planning and action modules \citep{aboualiAgenticAIComprehensive2025}.

\section{Conclusions}

In this study, we present NeuroCBIR, a content-based image retrieval framework for brain MRI that supports both whole-brain and region-specific retrieval. The system leverages latent embeddings from a variational autoencoder combined with contrastive learning, allowing efficient and semantically meaningful representations of 3D T1w MRIs. 

NeuroCBIR was evaluated with \textgreater\num[round-mode=places,round-precision=0]{26000} MRI scans from five large-scale and complementary datasets (ADNI, OASIS3, AIBL, MIRIAD, and SLIM), encompassing a wide range of acquisition protocols, scanner manufacturers, and field strengths. The results consistently demonstrated a high retrieval performance: whole-brain embeddings achieved mAP@5~$\ge98.4\%$, while region-specific retrieval provided strong performance across most brain anatomical structures. The Spearman’s correlation between the embedding-space and MS-SSIM ($\rho_{\mathrm{MS-SSIM}}$) further confirms that the learned embeddings preserve meaningful perceptual and semantic similarities between inter-subject scans. These findings confirm the robustness and generalization of the approach, even under heterogeneous imaging conditions. 

Additionally, the embeddings demonstrate to capture rich, subject-specific anatomical patterns that support subject re-identification in large datasets, zero-shot age prediction and multi-class pathology stratification without any supervised training. By leveraging a retrieval-based approach, our framework provides interpretable predictions and reveals meaningful disease-related structure, including accelerated aging patterns in AD and intermediate representations for MCI. Importantly, because the embeddings are fully unsupervised and task-agnostic, they can be readily applied to a wide range of downstream neuroscience applications, offering a flexible and interpretable alternative to traditional supervised models while reducing reliance on large labeled datasets.

The embedding extraction on a 4-core CPU required $\sim$18.7s per scan, whereas similarity search was virtually instantaneous (\textless0.01s). By making the code publicly available, the pretrained models and the precomputed features of \textgreater\num[round-mode=places,round-precision=0]{26000} from five datasets, NeuroCBIR enables clinicians to retrieve structurally similar cases to support diagnosis, monitoring, and treatment decisions. For the research community, NeuroCBIR provides a reproducible platform for benchmarking and extending CBIR approaches in neuroimaging.

\section*{CRediT authorship contribution statement}

Félix Nieto-del-Amor (FNA): Conceptualization, Methodology, Software, Formal analysis, Investigation, Visualization, Writing – original draft, Writing – review \& editing, Data curation.
Jingru Fu (JF): Methodology, Writing – review \& editing. 
J-Sebastian Muehlboeck (SM): Methodology, Writing – review \& editing, Data curation. 
Eric Westman (EW): Methodology, Writing – review \& editing. 
Daniel Ferreira (DF): Methodology, Writing – review \& editing. 
Rodrigo Moreno (RM): Conceptualization, Methodology, Supervision, Resources, Funding acquisition, Writing – review \& editing, Project administration.

\section*{Funding}

This work was supported in part by a postdoctoral fellowship from Digital Futures awarded to Félix Nieto-del-Amor (KTH-RPROJ-0146472). The project was also partially funded by the Swedish Research Council grant 2022-03389, Digital Futures and MedTechLabs. D. Ferreira receives funding from the Swedish Research Council grant 2022-00916, the Center for Innovative Medicine (CIMED, grants 20200505 and FoUI-988826), the regional agreement on medical training and clinical research of Stockholm Region (ALF Medicine, grants FoUI-962240 and FoUI-987534), the Swedish Brain Foundation (Hjärnfonden FO2025-0214, FO2023-0261, FO2022-0175, FO2021-0131), the Swedish Alzheimer Foundation (Alzheimerfonden AF-968032, AF-980580, AF-994058, AF-1010553), the Swedish Dementia Foundation (Demensfonden), the Gamla Tjänarinnor Foundation, the Gun och Bertil Stohnes Foundation, the Åke Wiberg Foundation, the Strategic Research Programme in Neuroscience at Karolinska Institutet (StratNeuro) Bridging Grant, Parkinsonfonden, Funding for Research from Karolinska Institutet, Neurofonden, and the Foundation for Geriatric Diseases at Karolinska Institutet, contributions from private bequests and academic agreements with industry.


\section*{Data availability}
The datasets used in this paper are publicly available under the specific conditions of every data provider. \reviiadd{NeuroCBIR code and a list of all scans included from each dataset is available in our GitHub repository (\url{https://github.com/minnelab/NeuroCBIR})}.

\section*{Acknowledgements}
{\footnotesize
Data were provided in part by ADNI: 
    Data collection and sharing for this project was funded by the Alzheimer's Disease Neuroimaging Initiative (ADNI) (National Institutes of Health Grant U01 AG024904) and DOD ADNI (Department of Defense award number W81XWH-12-2-0012). ADNI is funded by the National Institute on Aging, the National Institute of Biomedical Imaging and Bioengineering, and through generous contributions from the following: AbbVie, Alzheimer’s Association; Alzheimer’s Drug Discovery Foundation; Araclon Biotech; BioClinica, Inc.; Biogen; Bristol-Myers Squibb Company; CereSpir, Inc.; Cogstate; Eisai Inc.; Elan Pharmaceuticals, Inc.; Eli Lilly and Company; EuroImmun; F. Hoffmann-La Roche Ltd and its affiliated company Genentech, Inc.; Fujirebio; GE Healthcare; IXICO Ltd.; Janssen Alzheimer Immunotherapy Research \& Development, LLC.; Johnson \& Johnson Pharmaceutical Research \& Development LLC.; Lumosity; Lundbeck; Merck \& Co., Inc.; Meso Scale Diagnostics, LLC.; NeuroRx Research; Neurotrack Technologies; Novartis Pharmaceuticals Corporation; Pfizer Inc.; Piramal Imaging; Servier; Takeda Pharmaceutical Company; and Transition Therapeutics. The Canadian Institutes of Health Research is providing funds to support ADNI clinical sites in Canada. Private sector contributions are facilitated by the Foundation for the National Institutes of Health (\url{www.fnih.org}). The grantee organization is the Northern California Institute for Research and Education, and the study is coordinated by the Alzheimer’s Therapeutic Research Institute at the University of Southern California. ADNI data are disseminated by the Laboratory for Neuro Imaging at the University of Southern California. 

Data were provided in part by OASIS3:
    (i) OASIS-3: Longitudinal Multimodal Neuroimaging: Principal Investigators: T. Benzinger, D. Marcus, J. Morris; NIH P30 AG066444, P50 AG00561, P30 NS09857781, P01 AG026276, P01 AG003991, R01 AG043434, UL1 TR000448, R01 EB009352. AV-45 doses were provided by Avid Radiopharmaceuticals, a wholly owned subsidiary of Eli Lilly. (ii) OASIS-3\_AV1451: Principal Investigators: T. Benzinger, J. Morris; NIH P30 AG066444, AW00006993. AV-1451 doses were provided by Avid Radiopharmaceuticals, a wholly owned subsidiary of Eli Lilly.

Data were provided in part by MIRIAD:
    Data used in the preparation of this article were obtained from the MIRIAD database (\url{http://miriad.drc.ion.ucl.ac.uk}). The MIRIAD investigators did not participate in analysis or writing of this report. The MIRIAD dataset is made available through the support of the UK Alzheimer's Society (Grant RF116). The original data collection was funded through an unrestricted educational grant from GlaxoSmithKline (Grant 6GKC).
    
Data were provided in part by SLIM:
    We gratefully acknowledge the use of the SLIM (Southwest University Longitudinal Imaging Multimodal) dataset (available at \url{https://fcon_1000.projects.nitrc.org/indi/retro/southwestuni_qiu_index.html}). SLIM provides healthy, young adult test–retest neuroimaging data, including structural MRI, diffusion‐weighted MRI, resting‐state fMRI, behavioral and demographic measures over multiple timepoints. We thank the investigators and participants of the SLIM project for making these data publicly available.
    
Data were provided in part by AIBL: 
    Data used in the preparation of this article was obtained from the Australian Imaging Biomarkers and Lifestyle flagship study of ageing (AIBL). See \url{www.aibl.csiro.au} for further details.

This study used the \textit{TheHiveDB} from Karolinska Institutet for data collection and preprocessing. We gratefully acknowledge the developers and maintainers of TheHiveDB for providing this resource.

The computations/data handling were enabled by resources provided by the National Academic Infrastructure for Supercomputing in Sweden (NAISS), partially funded by the Swedish Research Council through grant agreement no. 2022-06725. The computations were further enabled by the Berzelius resource provided by the Knut and Alice Wallenberg Foundation. We also used the MAIA platform at KTH \citep{Bendazzoli2025} for the computations.

}

\section*{Declaration of generative AI and AI-assisted technologies in the manuscript preparation process}

During the preparation of this work, the authors used \mbox{ChatGPT} to enhance the clarity, coherence, and overall quality of the writing. After using this tool/service, the authors reviewed and edited the content as needed and take full responsibility for the content of the published article.

\bibliographystyle{elsarticle-harv} 
\bibliography{refs}

\appendix
\section{Detailed Brain-Region Metrics}
\label{app:detailed_region_metrics}

This appendix presents the full brain-region retrieval results of the NeuroCBIR framework. Table~\ref{tab:detailed_region_metrics} reports mAP@5, mS@1, and mS@5 for all evaluated regions across the testing partition, providing a detailed view of the system’s performance at the level of individual anatomical structures.

\begin{table*}[!ht]
\begin{threeparttable}
\caption{mAP@5, mS@1 and mS@5 for different brain regions of the NeuroCBIR framework across the testing dataset.}
\label{tab:detailed_region_metrics}
\centering
\footnotesize
\begin{tabular*}{\textwidth}{@{\extracolsep{\fill}} p{3.8cm} c c c p{3.8cm} c c c @{}}
\toprule
\textbf{Structure} & \reviadd{\textbf{mAP@5}} & \reviadd{\textbf{mS@1}} & \reviadd{\textbf{mS@5}} 
& \textbf{Structure} & \reviadd{\textbf{mAP@5}} & \reviadd{\textbf{mS@1}} & \reviadd{\textbf{mS@5}} \\
\midrule
Right-Cerebral-White-Matter & 99.6 & 99.4 & 100.0 & Left-Cerebellum-White-Matter & 98.3 & 96.8 & 99.9 \\
Left-Cerebral-White-Matter & 99.5 & 99.4 & 99.9 & ctx-lh-lateralorbitofrontal & 98.2 & 98.3 & 99.6 \\
Right-Cerebellum-Cortex & 99.5 & 98.6 & 99.9 & ctx-lh-pericalcarine & 98.2 & 97.8 & 99.4 \\
ctx-lh-superiortemporal & 99.5 & 99.3 & 99.9 & ctx-rh-parstriangularis & 98.1 & 98.0 & 99.6 \\
ctx-rh-fusiform & 99.5 & 99.0 & 99.9 & ctx-lh-bankssts & 98.1 & 99.0 & 99.6 \\
ctx-lh-fusiform & 99.5 & 98.8 & 99.9 & ctx-rh-bankssts & 98.0 & 99.2 & 99.7 \\
ctx-rh-superiortemporal & 99.4 & 99.4 & 100.0 & ctx-lh-transversetemporal & 98.0 & 97.6 & 99.8 \\
ctx-lh-precuneus & 99.4 & 98.9 & 99.9 & Right-Cerebellum-White-Matter & 98.0 & 97.2 & 99.8 \\
ctx-rh-middletemporal & 99.4 & 99.4 & 99.9 & ctx-rh-paracentral & 98.0 & 98.5 & 99.7 \\
ctx-lh-inferiortemporal & 99.3 & 99.0 & 99.8 & ctx-rh-lateralorbitofrontal & 97.9 & 97.7 & 99.4 \\
Brain-Stem & 99.3 & 98.9 & 100.0 & ctx-rh-parsorbitalis & 97.4 & 97.2 & 99.4 \\
Right-Lateral-Ventricle & 99.3 & 98.9 & 99.9 & ctx-lh-parsorbitalis & 97.3 & 97.2 & 99.4 \\
ctx-lh-superiorfrontal & 99.3 & 99.0 & 99.9 & ctx-rh-caudalanteriorcingulate & 97.2 & 97.1 & 99.7 \\
Left-Cerebellum-Cortex & 99.3 & 98.6 & 99.9 & ctx-rh-medialorbitofrontal & 97.2 & 97.2 & 99.4 \\
Left-Lateral-Ventricle & 99.3 & 98.8 & 99.9 & ctx-lh-rostralanteriorcingulate & 97.1 & 98.0 & 99.6 \\
ctx-rh-inferiortemporal & 99.3 & 98.9 & 99.7 & ctx-lh-caudalanteriorcingulate & 97.0 & 97.0 & 99.6 \\
ctx-lh-middletemporal & 99.3 & 99.1 & 99.8 & ctx-lh-medialorbitofrontal & 97.0 & 97.6 & 99.4 \\
ctx-lh-postcentral & 99.2 & 99.2 & 99.9 & ctx-rh-posteriorcingulate & 97.0 & 98.0 & 99.6 \\
ctx-rh-lingual & 99.2 & 98.8 & 99.8 & ctx-lh-posteriorcingulate & 96.9 & 98.1 & 99.7 \\
ctx-lh-precentral & 99.2 & 99.3 & 99.9 & ctx-rh-transversetemporal & 96.7 & 97.1 & 99.6 \\
ctx-lh-parsopercularis & 99.2 & 98.9 & 99.9 & ctx-rh-rostralanteriorcingulate & 96.6 & 97.0 & 99.4 \\
ctx-rh-precentral & 99.2 & 99.1 & 99.9 & Left-VentralDC & 96.6 & 95.5 & 99.6 \\
ctx-lh-lingual & 99.2 & 98.7 & 99.8 & Right-VentralDC & 96.5 & 94.9 & 99.7 \\
ctx-rh-precuneus & 99.2 & 99.2 & 99.8 & WM-hypointensities & 95.4 & 94.6 & 99.4 \\
ctx-rh-superiorfrontal & 99.2 & 99.0 & 99.8 & 4th-Ventricle & 95.1 & 94.9 & 99.6 \\
ctx-lh-superiorparietal & 99.2 & 98.9 & 99.8 & Right-Inf-Lat-Vent & 94.7 & 93.7 & 99.1 \\
ctx-lh-supramarginal & 99.2 & 99.0 & 99.8 & 3rd-Ventricle & 94.6 & 89.2 & 99.6 \\
ctx-rh-supramarginal & 99.2 & 99.1 & 99.8 & Left-Inf-Lat-Vent & 94.0 & 93.5 & 99.2 \\
ctx-lh-caudalmiddlefrontal & 99.1 & 99.1 & 100.0 & Right-Thalamus & 93.8 & 92.2 & 99.5 \\
ctx-lh-insula & 99.1 & 98.8 & 99.8 & Left-Thalamus & 93.6 & 91.5 & 99.4 \\
ctx-rh-inferiorparietal & 99.1 & 99.1 & 99.8 & ctx-rh-frontalpole & 92.3 & 94.9 & 98.1 \\
ctx-lh-rostralmiddlefrontal & 99.1 & 98.9 & 99.7 & Left-Caudate & 92.3 & 93.3 & 99.0 \\
ctx-lh-lateraloccipital & 99.1 & 98.3 & 99.6 & ctx-lh-frontalpole & 92.0 & 94.8 & 98.4 \\
ctx-lh-inferiorparietal & 99.0 & 98.7 & 99.7 & Right-Caudate & 91.2 & 92.4 & 99.3 \\
ctx-rh-postcentral & 99.0 & 99.2 & 99.8 & ctx-lh-entorhinal & 89.9 & 92.6 & 98.6 \\
ctx-rh-rostralmiddlefrontal & 99.0 & 98.2 & 99.8 & Right-Putamen & 89.7 & 91.9 & 98.8 \\
ctx-rh-superiorparietal & 99.0 & 98.9 & 99.8 & ctx-rh-entorhinal & 89.2 & 92.1 & 98.7 \\
ctx-rh-parsopercularis & 99.0 & 98.7 & 99.8 & Left-Putamen & 88.9 & 91.2 & 98.5 \\
ctx-rh-insula & 99.0 & 98.7 & 99.8 & ctx-rh-temporalpole & 86.8 & 91.9 & 96.6 \\
ctx-lh-isthmuscingulate & 98.9 & 98.5 & 99.9 & CC\_Anterior & 86.0 & 93.5 & 98.4 \\
ctx-lh-cuneus & 98.9 & 98.5 & 99.6 & ctx-lh-temporalpole & 85.3 & 90.8 & 95.8 \\
ctx-rh-caudalmiddlefrontal & 98.9 & 98.6 & 99.8 & CC\_Posterior & 81.4 & 88.8 & 94.1 \\
ctx-rh-lateraloccipital & 98.8 & 98.6 & 99.6 & Right-Amygdala & 80.2 & 83.9 & 98.0 \\
ctx-lh-parahippocampal & 98.7 & 98.1 & 99.7 & Left-Amygdala & 75.5 & 80.5 & 96.3 \\
ctx-rh-isthmuscingulate & 98.7 & 98.7 & 99.9 & CC\_Mid\_Posterior & 72.2 & 84.3 & 94.3 \\
ctx-lh-parstriangularis & 98.7 & 98.7 & 99.8 & CC\_Mid\_Anterior & 70.6 & 84.1 & 95.6 \\
ctx-rh-cuneus & 98.7 & 98.6 & 99.6 & CC\_Central & 62.1 & 77.8 & 92.4 \\
ctx-rh-parahippocampal & 98.5 & 98.2 & 99.9 & Left-Pallidum & 61.1 & 68.8 & 91.5 \\
ctx-rh-pericalcarine & 98.3 & 98.1 & 99.5 & Right-Pallidum & 58.3 & 66.6 & 90.2 \\
Left-Hippocampus & 98.3 & 97.7 & 99.8 & Right-Accumbens-area & 50.6 & 60.8 & 87.4 \\
ctx-lh-paracentral & 98.3 & 98.4 & 99.7 & Left-Accumbens-area & 40.8 & 52.5 & 82.6 \\
Right-Hippocampus & 98.3 & 98.1 & 99.8 &  &  &  &  \\
\bottomrule
\end{tabular*}

\begin{tablenotes}
        {\footnotesize
        \setlength{\leftmargin}{0pt}
        \setlength{\labelsep}{0pt}
        \item[] Note: The abbreviations used in the table are as follows: CC = Corpus Callosum; Mid = Middle; lh/rh = left/right
hemisphere; ctx = cortex.}
     \end{tablenotes}
\end{threeparttable}
\end{table*}




\end{document}